\documentclass{article}

\usepackage[main, final]{neurips_2026}

\usepackage[utf8]{inputenc} 
\usepackage[T1]{fontenc}    
\usepackage{hyperref}       
\usepackage{url}            
\usepackage{booktabs}       
\usepackage{amsfonts}       
\usepackage{nicefrac}       
\usepackage{microtype}      
\usepackage{xcolor}         
\usepackage{marvosym}

\newcommand{\TODO}[1]{\textbf{\color{red}[TODO: #1]}}
\renewcommand{\TODO}[1]{}

\usepackage{graphicx}
\usepackage{multirow}
\usepackage{xcolor}
\usepackage{booktabs}
\usepackage{adjustbox}
\usepackage{colortbl}
\usepackage{makecell}
\usepackage{tabularx}   

\usepackage{amsmath, amssymb}
\usepackage{algorithm}
\usepackage{algpseudocode}
\usepackage{bm}      
\usepackage{ulem}    

\usepackage{bbding}
\usepackage{utfsym}
\usepackage{pifont}
\usepackage{fontawesome}
\usepackage{listings}
\usepackage{color}
\usepackage{caption}
\usepackage{wrapfig}

\usepackage[capitalize]{cleveref}
\crefname{section}{Sec.}{Secs.}
\Crefname{section}{Section}{Sections}
\Crefname{table}{Table}{Tables}
\crefname{table}{Tab.}{Tabs.}

\usepackage{algorithm}

\definecolor{codeblue}{rgb}{0.25,0.5,0.5}
\definecolor{codekw}{rgb}{0.85, 0.18, 0.50}
\definecolor{codegreen}{rgb}{0,0.6,0}
\definecolor{codegray}{rgb}{0.5,0.5,0.5}
\definecolor{codepurple}{rgb}{0.58,0,0.82}
\definecolor{backcolour}{rgb}{0.95,0.95,0.92}

\definecolor{coolblack}{rgb}{0.4, 0.01, 0.24}

\lstdefinestyle{mystyle}{
    commentstyle=\color{codegreen},
    keywordstyle=\color{magenta},
    numberstyle=\tiny\color{codegray},
    stringstyle=\color{codepurple},
    basicstyle=\ttfamily\scriptsize,
    breakatwhitespace=false,
    breaklines=true,
    captionpos=b,
    keepspaces=true,
    numbers=left,
    numbersep=5pt,
    showspaces=false,
    showstringspaces=false,
    showtabs=false,
    tabsize=2,
    language=Python 
}
\usepackage{amsmath,amsfonts,bm}

\newcommand{\x}{\vx}

\newcommand{\veps}{\bm{\varepsilon}}

\def\eqref#1{equation~\ref{#1}}

\def\1{\bm{1}}

\def\ve{{\bm{e}}}

\def\vh{{\bm{h}}}

\def\vs{{\bm{s}}}

\def\vx{{\bm{x}}}

\def\mM{{\bm{M}}}

\DeclareMathAlphabet{\mathsfit}{\encodingdefault}{\sfdefault}{m}{sl}
\SetMathAlphabet{\mathsfit}{bold}{\encodingdefault}{\sfdefault}{bx}{n}

\def\gD{{\mathcal{D}}}
\def\gE{{\mathcal{E}}}

\def\gL{{\mathcal{L}}}

\newcommand{\E}{\mathbb{E}}

\title{VUM: Visual Unified Model for Image Generation and Perception}

\author{
    \normalfont
    \centerline{
        Zidong Wang$^{1}$\textsuperscript{*}\textsuperscript{\Letter}
        \quad
        Yiyuan Zhang$^{1}$\textsuperscript{*}\textsuperscript{\Letter}
        \quad
        Xiaoyu Yue$^{2}$ \quad
        Xiangyuan Xue$^{1}$ 
    } 
    \\
    \centerline{
        Zhangquan Chen$^{3}$ \quad
        Manyuan Zhang$^{1}$\textsuperscript{\dag} \quad
        Wanli Ouyang$^{1}$\textsuperscript{\dag} 
    } 
    \\
    \\
    \centerline{
        $^{1}$The Chinese University of Hong Kong \quad
        $^{2}$The University of Sydney
        \quad 
        $^{3}$Tsinghua University
    } 
}

\begin{document}

\maketitle

\begingroup
\renewcommand{\thefootnote}{}
\footnotetext[0]{\textsuperscript{*}Equal contribution.\quad \textsuperscript{\dag}Corresponding authors.\quad \textsuperscript{\Letter}Email: \{wangzd2022, yiyuanzhang.ai\}@gmail.com.}
\endgroup

\begin{figure*}[htbp]
\begin{center}
\centerline{\includegraphics[width=1.0\linewidth]{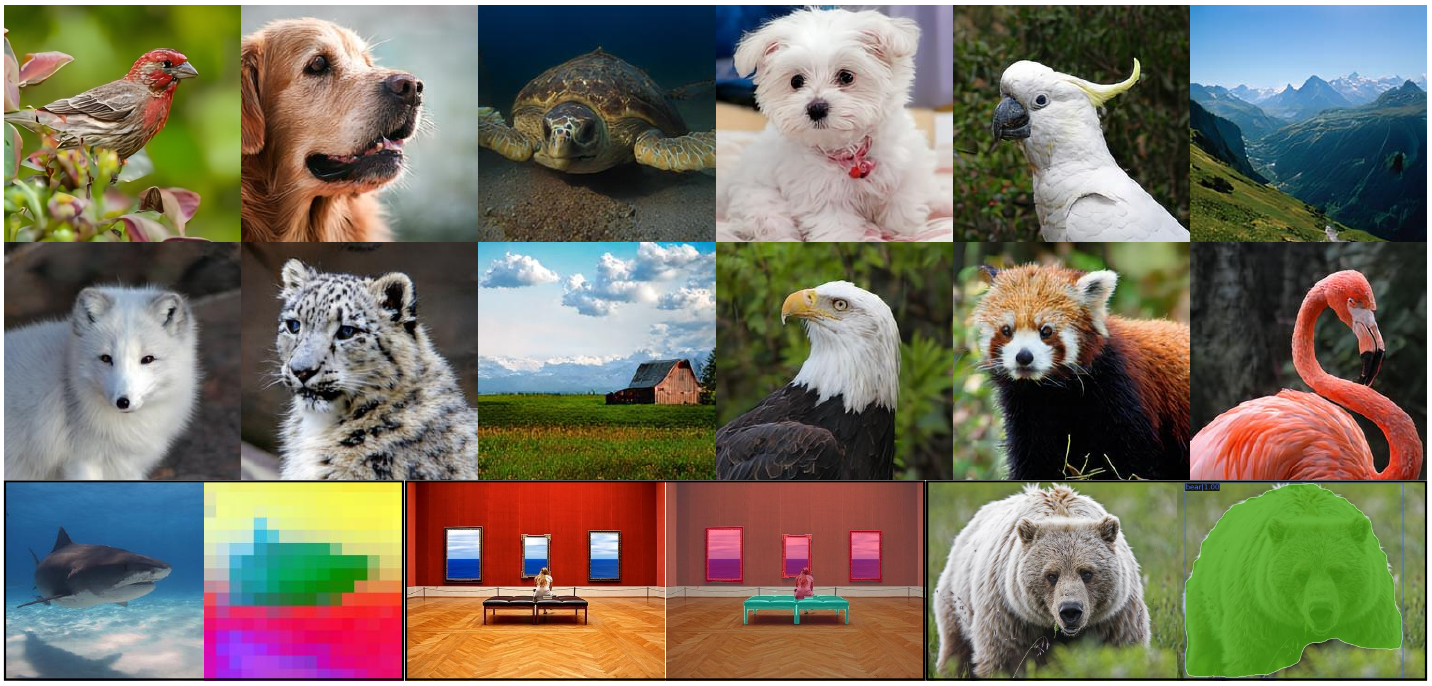}}
\caption{\textbf{One unified model, dual capabilities.} The Visual Unified Model (VUM) seamlessly integrates pixel synthesis and semantic perception within a single architecture. The top two rows showcase high-fidelity image generation on the ImageNet $256\times256$ benchmark, while the bottom row demonstrates its ability to extract highly effective representations for dense understanding tasks, including object detection, semantic segmentation, and instance segmentation.}
\label{fig:comparison}
\end{center}
\end{figure*}

\begin{abstract}

Unifying visual perception and generation poses a fundamental dichotomy: perception extracts semantics but discards details, while generation synthesizes fine-grained structures. Thus, existing works rely on decoupled architectures, as forcing these opposing information flows into one network inevitably triggers task interference and representation collapse. 
To break this bottleneck, we introduce the Visual Unified Model (VUM), a native framework that unifies both capabilities within a single architecture. Our key insight is that visual perception and generation can be unified by re-envisioning Masked Image Modeling and Diffusion Processes as complementary facets of data recovery. 
By establishing a shared degradation space that integrates both masking and noising, VUM optimizes a joint signal recovery objective: simultaneously predicting masked patches and denoising corrupted signals. This objective enforces the dual internalization of semantic abstraction and generative synthesis. Using one shared encoder, VUM achieves competitive image generation and transfers to visual recognition and dense prediction within a single pixel-native model.
Our implementation code is available at: \href{https://github.com/WZDTHU/VUM}{https://github.com/WZDTHU/VUM}.

\end{abstract}

\section{Introduction}



Unifying perception and generation within a single foundation model is a critical milestone toward Artificial General Intelligence (AGI)~\cite{hurst2024gpt4o,singh2025gpt5,wang2024emu3,comanici2025gemini2.5}.
Large Language Models (LLMs)~\cite{meta2025llama4,yang2024qwen2.5,brown2020gpt3,team2025gemma3} achieve this through the autoregressive ``next-token prediction'' paradigm. The causal nature of text compels the model to comprehend the preceding context as a prerequisite for synthesizing future content. 
In computer vision, however, perception and generation remain bifurcated.  Diffusion  models~\cite{ho2020ddpm,peebles2023dit,Lu2024FiT,wang2024fitv2,esser2024sd3,podell2023sdxl,wei2026unipic3} optimize low-level sensory fidelity without incentivizing semantic abstraction, whereas self-supervised learning~\cite{oquab2023dinov2,He2019moco,Chen2020simclr} extracts invariant semantics via lossy compression, discarding the spatial granularity crucial for generation. Consequently, visual capabilities remain isolated across sensory and semantic spaces. Lacking the inherent causality of language, vision still misses an analogous mechanism to bridge this gap under a singular optimization signal.


Recent latent-space approaches~\cite{zheng2025rae,shi2025svg,sun2023emu}, attempt to circumvent this divide through a multi-stage pipeline anchored to pre-trained vision foundation models (VFM)\cite{He2021mae,oquab2023dinov2,He2019moco,radford2021clip}. A frozen VFM first projects visual signals into latent tokens; a VAE decoder is then trained for pixel reconstruction on these tokens~\cite{rombach2022ldm,esser2021taming}; finally, a generative backbone (e.g., DiT~\cite{peebles2023dit}) is independently trained for synthesis. As illustrated in \cref{fig:comparison}, this design delegates \textit{perception} entirely to the vision encoder and \textit{generation} to the generative backbone. Consequently, these capabilities remain structurally decoupled; rather than being unified within a single learning objective, they are merely stitched together and forced to communicate through the artificial bottleneck of latent tokens.

\begin{figure*}[t]
\begin{center}
\centerline{\includegraphics[width=1.0\linewidth]{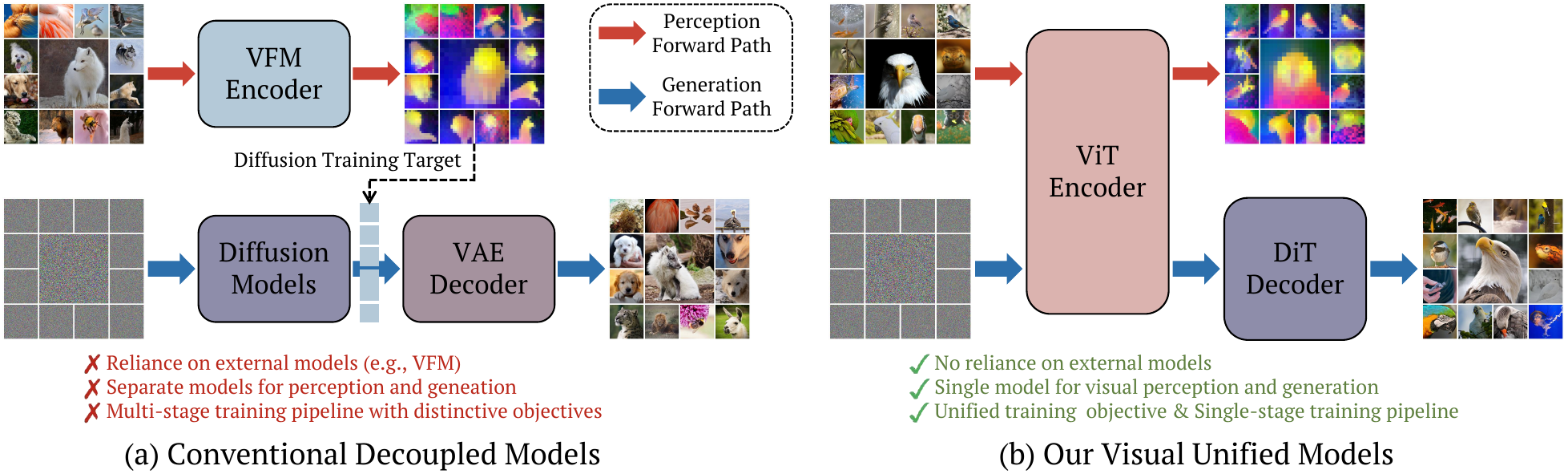}}
\caption{\textbf{Comparison of visual unification modeling paradigms.} (a) Conventional methods~\cite{zheng2025rae,shi2025svg} rely on an external Visual Foundation Model (VFM) as a feature extractor, employing multi-stage pipelines that require separate VAE and diffusion model training for synthesis. (b) Our VUM integrates perception and generation into a single backbone, enabling a single-stage training pipeline.}
\label{fig:comparison}
\end{center}
\end{figure*}

A fundamental question remains unresolved: \textit{What constitutes the true native paradigm for unifying visual perception and generation?} We draw our initial inspiration from the structural isomorphism between diffusion formulation~\cite{ho2020ddpm,lipman2022flowmatching,song2020sde} and masked image modeling (MIM)~\cite{He2021mae,Xie2021simmim,bao2021beit}. Despite driving distinct capabilities historically, both paradigms share a fundamental objective: recovering the original data manifold from corrupted observations, whether through continuous Gaussian noise or discrete spatial masking.
This observation motivates our core insight: \textit{the paradigm for visual unification is not a multi-task ensemble, but a singular, joint signal recovery process.} Rather than treating discrete masking and continuous noising as isolated tasks, we advocate for a joint visual signal recovery objective: the model must simultaneously deduce masked patches from visible context and reconstruct clean signals from Gaussian perturbations. This dual constraint fundamentally dictates the network's evolution: masked reconstruction forces the extrapolation of global semantics, while denoising compels the synthesis of high-fidelity structures. Consequently, a single network internalizes both perceptual abstraction and generative synthesis.

However, actualizing this joint paradigm is non-trivial. Naively superimposing discrete masking and continuous noising is highly ineffective. 
This approach not only induces severe information loss through compounded degradation but also triggers gradient interference within a shared architecture, ultimately resulting in catastrophic mutual degradation that severely compromises both capabilities.


To this end, we introduce the Visual Unified Model (VUM), a native framework that unifies visual generation and understanding within a singular architecture through three key innovations:
\textbf{1}) \textit{Unified Pixel-Level Pretraining via Information-Balanced Degradation}. 
By analyzing degradation through Signal-to-Noise Ratio (SNR)~\cite{oppenheim1997signals-and-systems}, we derive a Hybrid SNR (HSNR) formulation that mathematically bridges spatial masking and Gaussian noising. This metric precisely quantifies aggregate information loss, allowing us to dynamically couple the mask ratio with the noise level to ensure information-balanced degradation. By optimizing pixel-level signal recovery within this space, VUM captures both global semantic priors and fine-grained structural details.
\textbf{2}) \textit{Asymmetric Architecture Design}.
To resolve the inherent gradient conflicts during joint training, VUM employs a shared ViT~\cite{dosovitskiy2020vit} encoder coupled with an asymmetric dual-decoder. By pairing a primary DiT~\cite{peebles2023dit} decoder for denoising with an auxiliary ViT decoder for masked reconstruction, this design effectively mitigates optimization interference between semantics and generation.
\textbf{3}) \textit{Dual-Task Inference under a Singular Framework}. 
During inference, the auxiliary ViT decoder is discarded, allowing VUM to achieve bimodal execution without extraneous overhead. The ViT encoder extracts representations independently for perception while coupling with the DiT decoder for pixel synthesis. 
Thus, VUM unifies bimodal execution within a shared parameter space, bypassing the need for task-specific routing or network modifications. 

Extensive evaluations on image generation and downstream perception benchmarks validate the effectiveness of our design. To the best of our knowledge, VUM is the first framework to natively unify semantic perception and generative synthesis directly in pure pixel space. A single VUM achieves competitive pixel-space generative fidelity (an FID of 1.91 on the ImageNet $256\times256$ benchmark~\cite{deng2009imagenet}), while its shared encoder transfers to recognition and dense prediction, attaining 51.3 mIoU on ADE20K~\cite{zhou2019ade20k}, and 53.1 AP$^\text{box}$ and 46.3 AP$^\text{mask}$ on COCO~\cite{lin2014mscoco}. These results demonstrate competitive unification of pixel synthesis and semantic perception within a single pixel-native model.

\section{Related Work}

\noindent\textbf{Explorations of Visual Perception and Generation.} Current explorations predominantly rely on multi-stage pipelines, where an autoencoder \footnote{An autoencoder~\cite{hinton1993autoencoders,kingma2013vae} typically comprises an encoder and a decoder: the encoder compresses images into a latent space, while the decoder reconstructs the original pixels from these latent tokens.} is first optimized for feature extraction and reconstruction, followed by the training of an independent generative model within this latent space. To enhance perception, one line, such as VA-VAE~\cite{yao2025lightningdit}, VILA-U~\cite{wu2024vila-u}, and MAETok~\cite{chen2025maetok}, aligns the latent space with pre-trained vision models (e.g., DINO~\cite{caron2021dino,oquab2023dinov2}). Alternatively, frameworks like RAE~\cite{zheng2025rae} and SVG~\cite{shi2025svg} directly employ frozen pre-trained encoders, optimizing only the decoder for reconstruction. Others (UniTok~\cite{ma2025unitok}, VTP~\cite{yao2025vtp}) leverage cross-modal supervision; for instance, VTP incorporates textual modalities and the CLIP~\cite{radford2021clip} objective to augment visual perception.

Despite these advances, such methods share two limitations: 1) reliance on external priors (e.g., pre-trained vision features or text guidance), preventing true unification from raw pixels; and 2) structural decoupling, where perception and generation are handled by separate networks. Even concurrent joint-modeling efforts such as DREAM~\cite{li2026dream} still necessitate an auxiliary CLIP encoder to improve semantics. In contrast, VUM discards all external dependencies, coupling perception and generation within a single network through pure pixel-level degradation recovery.

\noindent\textbf{Self-Supervised Learning.} 
Advances in visual SSL can be broadly categorized into discriminative learning and masked image modeling (MIM) paradigms. Discriminative models (e.g., MoCo~\cite{He2019moco}, SimCLR~\cite{Chen2020simclr}, DINO~\cite{caron2021dino}) learn robust representations by maximizing agreement across augmented views. While this yields strong semantic discriminability for recognition tasks, it inevitably discards the fine-grained spatial details necessary for pixel-level synthesis.
Alternatively, MIM frameworks (e.g., MAE~\cite{He2021mae}, SimMIM~\cite{Xie2021simmim}, BEiT~\cite{bao2021beit}) capture rich contextual and perceptual information by reconstructing masked inputs to model the underlying data distribution. Although these representations trail behind discriminative methods in global semantic separability, their pixel-level reconstruction objective shares a critical structural isomorphism with diffusion denoising mechanisms. Motivated by this alignment, our VUM adopts the pixel-recovery paradigm to unify perception and synthesis.

\noindent\textbf{Pixel-Space Diffusion Models.} Despite the dominance of latent-space paradigms, direct modeling of pixels is resurging to preserve maximum signal fidelity. Early models (e.g., DDPM~\cite{ho2020ddpm}, ADM~\cite{dhariwal2021adm}) were anchored in convolutional U-Nets~\cite{ronneberger2015unet}; as the paradigm evolved to embrace Vision Transformers (ViTs), directly modeling raw pixels triggered a severe sequence length explosion at high resolutions. To mitigate this complexity, SiD2~\cite{Hoogeboom2025sid2} adopts a memory-efficient U-ViT~\cite{bao2023uvit} with level-wise skip connections, while PixelFlow~\cite{Chen2025PixelFlow} dynamically upscales samples as denoising progresses. Others utilize dual-level architectures to decouple global semantics from local refinement; for instance, PixNerd~\cite{wang2025pixnerd} conditions NeRF-like rendering~\cite{mildenhall2021nerf} on transformer features for patch-level reconstruction, and PixelDiT~\cite{Yu2025PixelDiT} cascades a patch-level DiT with a pixel-level DiT. 
In contrast, JiT~\cite{li2025jit} demonstrates that a plain ViT can directly predict clean images from raw patches without specialized architectural biases. Transcending this, our VUM establishes a singular, unified Transformer architecture that internalizes both perceptual understanding and generative synthesis.

\section{Method}

\subsection{Preliminaries}

\noindent\textbf{Masked Image Modeling.} 
As a canonical instantiation of Masked Image Modeling~\cite{bao2021beit,He2021mae,Xie2021simmim}, MAE~\cite{He2021mae} learns visual representations by reconstructing missing pixels from a set of visible tokens. Formally, given an image $\x\in\mathbb{R}^{H\times W\times 3}$, it is partitioned into $N$ non-overlapping patches. A binary mask $\mM\in\{0,1\}^{N}$ is sampled to mask $75\%$ of the patches, where $\mM_i=1$ indicates a masked patch. The encoder $\gE$ is a Vision Transformer (ViT) that operates exclusively on the sequence of visible tokens $\x_{\text{vis}}$ (extracted via $\x\odot\mM$). A generic learnable $\mathtt{[CLS]}$ token $\ve_{\text{cls}}$ is prepended to this sequence, yielding the latent representation $\vh=\gE([\ve_{\text{cls}}, \x_{\text{vis}}])$. The decoder $\gD$ processes the concatenated full sequence $\hat{\x}=\gD([\vh, \vs])$, where $\vs$ represents a set of shared and learned mask tokens placed at the missing positions ($\mM_i=1$). The loss is computed exclusively on the masked patches: $\gL_{\text{MAE}}=\E\|(\texttt{Norm}(\x)-\hat{\x})\odot\mM\|_2^2$, where $\texttt{Norm}(\cdot)$ denotes the unit-variance normalization.

\noindent\textbf{Diffusion Models.} 
As a prevalent formulation in diffusion models~\cite{ho2020ddpm,song2020sde,song2019ncsn,liu2023flow,lipman2022flowmatching,song2023cm,yue2024dod,wang2026nit,wang2026tim}, flow-matching~\cite{lipman2022flowmatching} linearly interpolates between sampled noise $\veps \sim \mathcal{N}(0,\mathbf{I})$ and data $\x \sim p_\text{data}(\x)$ via: $\x_t = (1-t) \x + t \veps$, where $t\in[0,1]$ is the sampled time step. In this scheduling, JiT~\cite{li2025jit} directly predicts the clean data manifold from noisy input: $\hat{\x}=\mathcal{D}_{\text{G}}(\x_t,t)$, where $\mathcal{D}_{\text{G}}$ is a Diffusion Transformer (DiT) architecture with a patch size of $16$. The training objective is to minimize the reconstruction error in the pixel space: $\gL_{\text{Diff}}=\E_t\|\x-\mathcal{D}_{\text{G}}(\x_t,t)\|^2_2$.

\subsection{Motivation}

To construct a truly unified visual model, an intuitive first step is to directly combine the objectives of MIM and diffusion within a single backbone. To empirically investigate the feasibility of this integration, we evaluate three baselines: (1) a standard MAE-B baseline; (2) a 24-layer JiT-B baseline \footnote{We scale the depth of JiT-B from the standard 12 layers to 24 layers to precisely match the total capacity of our VUM (detailed in \cref{eq:dynamic_r}) for a fair comparison. The three models are trained for 160 epochs on ImageNet-1K~\cite{deng2009imagenet}.}; and (3) a naively fused 24-layer JiT-B, trained concurrently with standard noising and $75\%$ spatial masking. We extract features from the $12$-th layer in all models for linear probing (IN1K-LP).

\begin{wraptable}{r}{0.34\linewidth}
    \vspace{-0.4cm}
    \begin{minipage}{\linewidth}
    \caption{\small Empirical baselines: Naïve integration of MIM and diffusion leads to catastrophic mutual degradation.}
    \label{tab:jit_fail}
    \centering
    \resizebox{\linewidth}{!}{
        \begin{tabular}{lcc}
        \toprule
        Config & FID$\downarrow$ & IN1K-LP$\uparrow$ \\
        \midrule
        MAE-B & - & 44.85 \\
        JiT-B (24-Layer) & 9.16  & 6.97  \\ 
        JiT-B (75$\%$ Mask)& 316.9 & 6.57  \\ 
        \bottomrule
        \end{tabular}
    }
    \end{minipage}
    \vspace{-0.4cm}
\end{wraptable}


\cref{tab:jit_fail} reveals three key findings: 1) pure MAE yields strong perceptual representations ($\sim 45\%$ accuracy) its asymmetric design and discrete reconstruction objective preclude conditional image generation; 2) pure JiT fails at understanding tasks ($<10\%$ IN1K-LP accuracy), demonstrating generative objectives alone lack perceptual abstraction; and 3) naively fusing both causes severe mutual degradation, deteriorating generation quality (FID 316.9) while keeping probing accuracy negligible.

These empirical failures indicate that perception and generation cannot be seamlessly unified by simply forcing a network to naively blend divergent training objectives. To establish a truly singular architecture, we must first dissect the deep-rooted incompatibilities that drive this mutual interference. Specifically, we identify four intrinsic challenges:
\textit{C1} \textbf{Compounding Information Loss.} Superimposing aggressive spatial masking (e.g., $75\%$) with diffusion's Gaussian noise severely depletes the input signal. Predicting pixels from such an information-starved state is fundamentally ill-posed, inevitably triggering optimization collapse.
\textit{C2} \textbf{Structural Asymmetry.} MAE favors an asymmetric, encoder-heavy architecture to abstract semantics, whereas generative models necessitate substantial decoder capacity for iterative spatial refinement.
\textit{C3} \textbf{Input Discrepancy.} During inference, perception operates on clean images, whereas diffusion synthesizes content from pure Gaussian noise under a specific condition. This creates a severe input manifold gap that a naive shared backbone struggles to bridge.
\textit{C4} \textbf{Misaligned Optimization Gradients.} The two objectives inherently conflict: MAE abstracts invariant semantics by discarding local details, while diffusion strictly retains high-frequency pixels. Consequently, a naive linear combination of these losses triggers severe gradient interference.

The structural and input discrepancies (\textit{C2} and \textit{C3}) are evident. Regarding Challenges 1 and 4, we provide further empirical evidence for the information degradation and optimization conflicts in the experiment \cref{sec:ablation}. Together, these limitations necessitate a rethinking of unified visual models.

\begin{figure*}[t]
\begin{center}
\centerline{\includegraphics[width=1.0\linewidth]{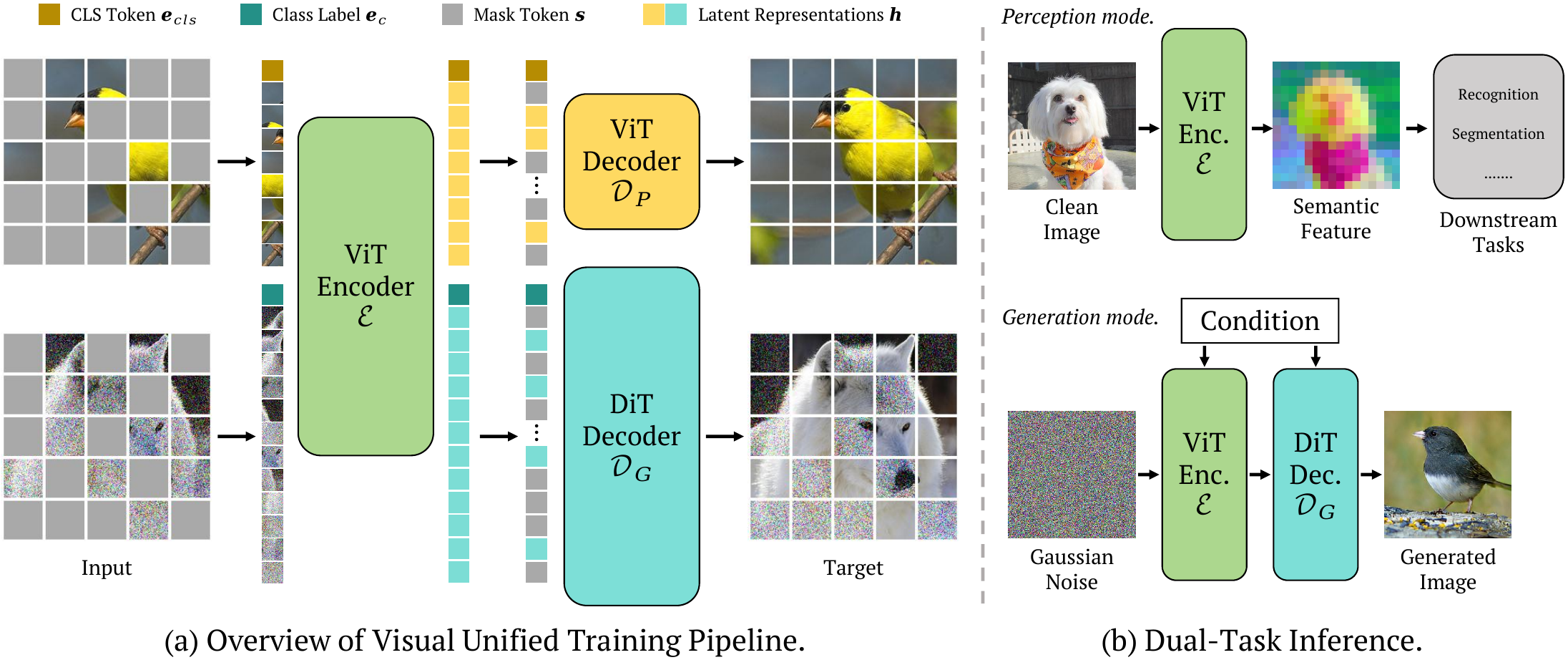}}
\caption{
\textbf{Overview of Visual Unified Models (VUM).} 
(a) Visual Unified Training. Inputs undergo joint degradation where the mask ratio dynamically adjusts with the noise level. A shared ViT encoder $\mathcal{E}$ extracts latent features, routed to task-specific decoders ($\mathcal{D}_P$, $\mathcal{D}_G$) for joint pixel-level recovery. 
(b) Dual-Task Inference. For perception, standalone $\mathcal{E}$ extracts semantics directly from clean images. For generation, $\mathcal{E}$ pairs with the DiT decoder $\mathcal{D}_G$ for conditional pixel synthesis from pure noise.
}
\label{fig:overview}
\end{center}
\vspace{-0.7cm}
\end{figure*}

\subsection{Visual Unified Models}


Overcoming the aforementioned barriers requires far more than a simple additive loss; it demands a paradigm shift from the degradation mechanism to the architectural design. As illustrated in \cref{fig:overview}, we present the Visual Unified Model (VUM), which dismantles the inherent perception-generation conflicts through three core designs:
\textit{1)} \textbf{Information-Balanced Degradation.} Directly addressing the issue of signal depletion (\textit{C1}), we introduce a joint scheduler governed by a Hybrid SNR formulation. By dynamically scaling the mask ratio inversely with the noise level, this formulation prevents excessive signal depletion.
\textit{2)} \textbf{Asymmetric Architecture with Semantic Bridging.} Addressing both structural mismatches (\textit{C2}) and input discrepancies (\textit{C3}), we design an asymmetric ViT-DiT topology. To bridge clean images and pure noise during inference, we dynamically repurpose the $\mathtt{[CLS]}$ token as a unified conditional anchor, seamlessly connecting the two modalities.
\textit{3)} \textbf{Dual-Decoder Design.} To eliminate the gradient interference typical of coupled architectures (\textit{C4}), VUM routes the latent features into decoupled task-specific heads. The generative synthesis is entrusted to a heavy DiT decoder, while a lightweight auxiliary ViT decoder handles pixel-level masked recovery.

\subsubsection{Information-Balanced Degradation}

\noindent\textbf{Hybrid Signal-to-Noise Ratio (HSNR).} To address the compounding information loss (\textit{C1}), we must quantify the aggregate signal destruction caused by concurrent masking and noising. To this end, we introduce the Hybrid Signal-to-Noise Ratio (HSNR), a metric that explicitly quantifies the ratio of preserved signal power to total visual uncertainty. We defer the rigorous mathematical derivations to Appendix \ref{app:snr_derivation} and summarize the core formulations here. First, the independent SNRs for continuous diffusion and discrete masking are formulated as:
\begin{equation}
    SNR_{\text{Diff}}(t) = \frac{(1-t)^2}{t^2}, \quad SNR_{\text{MAE}}(r) = \frac{1-r}{r}.
\end{equation}
Building upon these isolated states, we deduce the joint degradation. When superimposing a spatial mask $r$ onto a noisy continuous manifold $\mathbf{x}_t$, the preserved signal power becomes $(1-r)(1-t)^2$, while the total uncertainty aggregates the continuous noise $(1-r)t^2$ and the discrete spatial loss $r$. This mathematically yields the exact Hybrid SNR:
\begin{equation}
    SNR_{\text{Hybrid}}(r, t) = \frac{(1-r)(1-t)^2}{(1-r)t^2 + r}.
    \label{eq:snr_hybrid}
\end{equation}

\noindent\textbf{Diagnosis of Information Death and Dynamic Balance.} \cref{eq:snr_hybrid} formally exposes the root cause of the catastrophic collapse observed during naive fusion. A high mask ratio (e.g., $r=0.75$) inherently induces a massive baseline loss of visual information, which is then severely exacerbated as the noise level $t$ escalates. Specifically, as $t$ increases, the signal term $(1-t)^2$ in the numerator decays quadratically to zero, whereas the degradation term $(1-r)t^2 + r$ in the denominator strictly increases. This simultaneous signal decay and noise expansion causes the overall $SNR_{\text{Hybrid}}$ to non-linearly collapse toward zero (as shown in \cref{fig:snr_schedule}). Ultimately, this information death obliterates the visual signal, leaving the network with insufficient context to drive both generative and perceptual objectives.

To maintain a tractable information bottleneck throughout the forward process, we propose a dynamic \textit{Information-Balanced Scheduling} strategy. Rather than applying a static mask, we inversely couple the mask ratio $r$ with the continuous noise level $t$:
\begin{equation}
    r(t) = r_{\text{h}} - t \cdot (r_{\text{h}} - r_{\text{l}}),
    \label{eq:dynamic_r}
\end{equation}
where $r_{\text{h}} = 0.75$ and $r_{\text{l}} = 0.0$. By substituting this dynamic schedule back into \cref{eq:snr_hybrid}, we derive the Balanced HSNR, which is governed seamlessly by the temporal state $t$:
\begin{equation}
    SNR_{\text{Balanced}}(t) = \frac{(1 - r(t))(1-t)^2}{(1 - r(t))t^2 + r(t)}.
    \label{eq:snr_balanced}
\end{equation}
Crucially, this dynamic formulation transcends naive multi-task learning by orchestrating an elegant curriculum of signal recovery. Under severe continuous noise ($t \to 1$), the schedule naturally lifts the spatial mask ($r(t) \to 0$), equipping the network with the dense spatial context required for structural denoising. Conversely, as the signal clarifies ($t \to 0$), severe spatial masking is introduced ($r(t) \to 0.75$), forcing the network to extract high-level semantic abstractions rather than memorizing local pixels. Ultimately, perception and generation are no longer antagonistically coupled; instead, they operate as a unified, complementary continuum of signal recovery. We further provide empirical explorations of alternative scheduling functions for $r(t)$ in \cref{tab:ablations_schedule}.

\subsubsection{Architecture and Dual-Task Execution}

To resolve the remaining conflicts (\textit{C2–C4}), we introduce the architectural designs of Visual Unified Model (VUM). Addressing structural asymmetry (\textit{C2}), VUM pairs a shared ViT encoder with a heavy DiT decoder, directly satisfying perception's need for deep semantic abstraction and diffusion's demand for iterative spatial refinement. Building on this foundation, we further introduce specialized mechanisms to resolve input discrepancies (\textit{C3}) and mitigate gradient conflicts (\textit{C4}).


\noindent\textbf{Trajectory-Adaptive Context Modulation and Efficient Encoding.} To resolve the conditional input discrepancy (\textit{C3}) without fracturing the unified degradation continuum, we introduce a trajectory-adaptive context modulation mechanism. Rather than resorting to modality-specific branches, we formalize the standard $\mathtt{[CLS]}$ token slot as a \textit{noise-aware semantic pivot} governed by the trajectory $t \in [0, 1]$. As in \cref{fig:overview} (a), towards the clean-state limit ($t \to 0$), the slot is instantiated with a learnable embedding $\mathbf{e}_{\text{cls}}$ to aggregate global semantic abstraction. Across the noisy spectrum ($t > 0$), it incorporates the class condition $\mathbf{e}_c$ to guide the spatial synthesis. This context modulation mechanism seamlessly unifies feature extraction and conditional generation within the same encoder.


Furthermore, inspired by MAE~\cite{He2021mae}, the shared encoder $\gE$ operates exclusively on unmasked patches $\x_{\text{vis}}$ to maximize efficiency. Because our dynamic schedule $r(t)$ inherently creates highly variable sequence lengths within the same batch, we integrate a $\texttt{varlen}\_\texttt{flash}\_\texttt{attn}$ backend \footnote{Specifically, the $\texttt{varlen}\_\texttt{flash}\_\texttt{attn}$ interface~\cite{dao2023flashattention-2} packs disparate sequences into a 1D array, utilizing an index of cumulative lengths to enforce strict attention boundaries, achieving isolated self-attention without explicit masking overhead.}. This bypasses inefficient zero-padding and attention masking, allowing the encoder to project the modulated input sequences into a shared latent space $\vh = \gE([\mathbf{e}_{\text{cls/c}}, \x_{\text{vis}}])$ in a highly optimized forward pass.



\noindent\textbf{Decoupled Decoding as Gradient Buffers.} While the shared encoder unifies feature extraction, directly backpropagating from all temporal states causes severe gradient conflicts (further analyzed in the Appendix \ref{app:gradient_conflict}). To resolve this (\textit{C4}), we decouple the decoding phase into state-conditioned gradient buffers. After the encoder extracts the compressed latents $\vh$ and pads them with mask tokens $\vs$, the sequence is handled by two specialized decoders based on the trajectory state $t$:

\textit{Auxiliary ViT Decoder ($\gD_{\text{P}}$):} In the clean boundary state ($t=0$), a lightweight ViT (an $8$-layer ViT with a width of $512$) decoder forces the encoder to abstract semantics by recovering the missing pixels. It is supervised by a standard MAE loss computed exclusively on the masked regions:
\begin{equation}
    \gL_{\text{P}} = \E \left[ \| (\texttt{Norm}(\x) - \gD_{\text{P}}([\vh, \vs])) \odot \mM \|_2^2 \right].
    \label{eq:loss_perception}
\end{equation}

\textit{DiT Decoder ($\gD_{\text{G}}$):} A heavy generative decoder processes the continuous noisy stream ($t>0$). To handle noisy states with masking, $\gD_{\text{G}}$ employs a region-specific objective. It applies standard denoising to visible regions and reconstructs the noisy structure for masked regions:
\begin{equation}
    \gL_{\text{G}} = \E_{t} \left[ \| (\x - \gD_{\text{G}}([\vh, \vs], t)) \odot (1-\mM) \|_2^2 + \lambda \| (\texttt{Norm}(\x_t) - \gD_{\text{G}}([\vh, \vs], t)) \odot \mM \|_2^2 \right],
    \label{eq:loss_generation}
\end{equation}
where $\lambda$ balances the gradient scales between denoising and structure reconstruction, following \cite{zheng2023maskdit}. The full network is optimized end-to-end via a joint objective:
\begin{equation}
    \gL_{\text{Total}} = \beta \gL_{\text{P}} + \gL_{\text{G}}.
    \label{eq:loss_total}
\end{equation} 
Ultimately, driven by the joint objective $\gL_{\text{Total}}$, state-conditioned decoders act as independent gradient buffers: $\gD_{\text{P}}$ absorbs stable semantic gradients, while $\gD_{\text{G}}$ digests high-variance spatial dynamics. By resolving this divergence late in the network, VUM achieves a powerful synergy: it fully insulates the shared encoder to learn a universal visual abstraction, while simultaneously unlocking the DiT decoder to deliver high-fidelity generative synthesis without being bottlenecked by perceptual constraints.



\noindent\textbf{Dual-Task Execution.} As illustrated in \cref{fig:overview} (b), VUM executes perception and generation through an adaptive inference strategy. For perception, the encoder $\gE$ standalone extracts semantics for downstream adaptations. For generation, $\gE$ pairs with the DiT decoder $\gD_{\text{G}}$ to iteratively synthesize pixels from Gaussian noise, conditioned on $\mathbf{e}_c$. By discarding the auxiliary ViT decoder during inference, VUM maintains computational efficiency for both tasks without structural redundancy.

\section{Experiments}

\begin{table*}[t]
\caption{
\textbf{Model comparisons on ImageNet-256 Benchmark.} All models utilize Classifier-Free Guidance (CFG) for generation. Latent-space models incur an overhead of 86M parameters for the VAE~\cite{rombach2022ldm} or VQ-VAE~\cite{esser2021taming} tokenizer.
}
\label{tab:main_generation}
\vspace{-0.3cm}
\begin{center}
\begin{small}
\begin{adjustbox}{max width=\textwidth}
\begin{tabular}{l|lcc|ccccc}
\toprule
Type & Model & \#Params & Epochs & FID$\downarrow$ & sFID$\downarrow$ & IS$\uparrow$ & Prec.$\uparrow$ & Rec.$\uparrow$ \\
\midrule
\multirow{4}{*}{Autoregressive Models}
& VAR \cite{tian2024var} & 600M+86M & - & 2.57 & - & 302.6 & 0.83 & 0.56 \\
& LlamaGen \cite{sun2024autoregressive} & 775M+86M & 300 & 2.63 & 5.59 & 244.09 & 0.81 & 0.58 \\
& ST-AR \cite{yue2025st-ar} & 775M & 300 & 2.37 & 6.05 & 270.59 & 0.82 & 0.58 \\
& FractalMAR \cite{li2025fractalmar} & 848M & 600 & 6.15 & - & 348.9 & 0.81 & 0.46 \\
\midrule
\multirow{5}{*}{Latent Diffusion Models}
& LDM \cite{rombach2022ldm} & 400M+86M & 170 & 3.60 & 5.12 & 247.67 & 0.87 & 0.48  \\
& DiT \cite{peebles2023dit} & 675M+86M & 1400 & 2.27 & 4.60 & 278.24 & 0.83 & 0.57 \\
& MaskDiT \cite{zheng2023maskdit} & 700M+86M & 1600 & 2.28 & 5.67 & 276.60 & 0.80 & 0.61 \\
& SiT \cite{ma2024sit} & 675M+86M & 1400 & 2.15 & 4.50 & 258.09 & 0.81 & 0.60 \\
& REPA \cite{yu2024repa} & 675M+86M & 800 & 1.42 & 4.70 & 305.7 & 0.80 & 0.64 \\
\midrule
\multirow{10}{*}{Pixel Diffusion Models}
& CDM \cite{ho2022cdm} & - & 2160 & 4.88 & - & 158.7 & - & - \\
& ADM\cite{dhariwal2021adm} & 554M & 400 & 3.94 & 6.14 & 215.8 & 0.83 & 0.53 \\
& SiD \cite{hoogeboom2023simplediffusion} & 2.0B & 800 & 2.44 & - & 256.30 & - & -  \\
& PixelFlow \cite{Chen2025PixelFlow} & 677M & 320 & 1.98 & - & 282.1 & 0.81 & 0.60 \\
& DiP \cite{chen2025dip} & 631M & 320 & 1.98 & 4.57 & 282.9 & 0.80 & 0.60 \\
& PixNerd \cite{wang2025pixnerd} & 700M & 320 & 1.95 & - & 298.0 & 0.80 & 0.60 \\
& DeCo \cite{ma2511deco} & 682M & 320 & 1.90 & 4.47 & 303.0 & 0.80 & 0.61 \\
& JiT \cite{li2025jit} & 953M & 600 & 1.86 & - & 303.4 & 0.78 & 0.62 \\
& VUM (Ours) & 763M & 800 & 1.91 & 4.92 & 288.3 & 0.80 & 0.62 \\
\bottomrule
\end{tabular}
\end{adjustbox}
\end{small}
\end{center}
\vspace{-0.4cm}
\end{table*}

\begin{table*}[t]
\caption{\textbf{Model comparisons on downstream perception tasks.} All evaluated models employ a ViT-L backbone. For our method, we isolate the ViT encoder of VUM for downstream evaluation. We report mIoU for semantic segmentation on ADE20K, alongside AP$^\text{box}$ and AP$^\text{mask}$ for object detection and instance segmentation on COCO, respectively.}
\label{tab:main_perception}
\vspace{-0.3cm}
\begin{center}
\begin{small}
\begin{adjustbox}{max width=.85\textwidth}
\begin{tabular}{l|ll|cc|c|cc}
\toprule
\multirow{2}*{Model} & 
\multirow{2}*{Pretrain Data} & 
\multirow{2}*{Epoch.} & 
\multicolumn{2}{c|}{IN1K} & 
\multicolumn{1}{c|}{ADE20K} & 
\multicolumn{2}{c}{COCO}
\\
& & & FT$\uparrow$ & LP$\uparrow$ & mIOU$\uparrow$ & AP$^\text{box}\uparrow$ & AP$^\text{mask}\uparrow$ \\
\midrule
MoCo v3 \cite{chen2021mocov3}   & IN1K          & 600   & 84.1 & 77.6 & 49.1 & 49.3 & 44.0 \\
BEiT \cite{bao2021beit}     & IN1K+DALLE    & 1600  & 85.2 & - & 53.3 & 53.3 & 47.1 \\
MAE \cite{He2021mae}        & IN1K          & 1600  & 86.0 & 76.0 & 53.6 & 53.3 & 47.2 \\
VUM (Ours)                  & IN1K          & 800   & 84.4 & 68.4 & 51.3 & 53.1 & 46.3 \\
\bottomrule
\end{tabular}
\end{adjustbox}
\end{small}
\end{center}
\vspace{-0.5cm}
\end{table*}

\subsection{Setup}

\textbf{Implementation Setup.} The VUM architecture comprises a standard ViT encoder (patch size $16$) coupled with a DiT decoder and a ViT decoder. The flagship VUM-L configuration instantiates ViT-L and DiT-L backbones, yielding $763$M parameters.  Training is conducted on ImageNet at $256\times256$ resolution ($256$ patches per image) with standard center-crop and horizontal flip augmentations. 
The network is trained with Muon for $800$ epochs ($40$-epoch warmup), using a batch size of $2048$ and a learning rate of $2\times10^{-4}$. We set $\lambda=0.1$ in \cref{eq:loss_generation} and $\beta=1.0$ in \cref{eq:loss_total} by default. Following generative modeling conventions, an exponential moving average (EMA) with a $0.9999$ decay rate is maintained. More details are provided in the Appendix \ref{app:more_implementation_details}.  

\textbf{Evaluation Protocol.} Generative synthesis is evaluated using FID~\cite{heusel2017gans}, sFID~\cite{nash2021sfid}, Inception Score (IS)~\cite{salimans2017inceptionscore}, Precision and Recall~\cite{kynk2017precisionrecall}, standardizing the computation via the ADM suite~\cite{dhariwal2021adm}. Perception performance is evaluated through ImageNet-1K linear probing and end-to-end fine-tuning. We also evaluate VUM on downstream tasks: object detection and instance segmentation on COCO~\cite{lin2014mscoco} as well as semantic segmentation on ADE20K~\cite{zhou2019ade20k}.

\subsection{Unified Visual Capabilities on Generation and Perception}


\noindent\textbf{Image Generation Performance.} We evaluate the generative capabilities of VUM on the standard ImageNet-256 benchmark against state-of-the-art autoregressive, latent diffusion, and pixel-space diffusion models. As shown in \cref{tab:main_generation}, VUM achieves a competitive FID of 1.91 and an Inception Score (IS) of 288.3.
Within the pixel diffusion paradigm, VUM achieves an FID close to JiT and DeCo (1.91 vs. 1.86 and 1.90, respectively), while improving over ADM (3.94) and DiP (1.98).
Compared to latent diffusion models, VUM achieves lower FID and higher Recall than DiT (2.27) and MaskDiT (2.28) without relying on external VAEs. Together with the shared encoder's transfer performance, these results support competitive unification of generation and perception within a single pixel-native model.

\noindent\textbf{Perceptual Representation.} We evaluate the shared VUM encoder on downstream tasks in \cref{tab:main_perception}. For dense prediction, VUM achieves 51.3 mIoU on ADE20K and 53.1/46.3 AP on COCO, outperforming MoCo v3 and approaching MAE on COCO. VUM is pretrained for 800 epochs, compared with 1600 for MAE, while also learning image generation. On ImageNet-1K, VUM yields 84.4 FT accuracy and 68.4 LP accuracy, compared with 86.0 and 76.0 for MAE. The 7.6-percentage-point LP gap highlights a trade-off between generation and perceptual representation quality. Overall, the same encoder supports competitive transfer to recognition and dense prediction alongside image generation, with lower perceptual scores than MAE on these benchmarks.

\begin{table*}[t]
\caption{
\textbf{Ablations on the components of VUM.} Architectural configurations are denoted as \textit{Model-Layers/Width} (\emph{e.g.}, ViT-12/768 indicates a 12-layer ViT with a hidden dimension of 768).
$^\dag$: for the DiT-24/768 architectures, we use the 12-th layer feature for the end-to-end finetuning (IN1K-FT) and linear probing (IN1K-LP) evaluations on ImageNet1K. 
} 
\label{tab:ablations_component}
\vspace{-0.3cm}
\begin{center}
\begin{small}
\begin{adjustbox}{max width=\textwidth}
\begin{tabular}{l|l|ccc|c|cc|cc}
\toprule
Model & Params. & Encoder & Perceptual Decoder & Generative Decoder & Corruption & FID$\downarrow$ & IS$\uparrow$ & IN1K-FT$\uparrow$ & IN1K-LP$\uparrow$ \\
\midrule
\multicolumn{9}{l}{\textbf{\textit{Perception and Generation Baselines}}} \\
MAE & 148.7M & ViT-12/768 & ViT-8/512 & - & 75$\%$ Mask & - & - & 83.02 & 44.85 \\
JiT$^\dag$ & 258.7M & - & - & DiT-24/768 & Noise & 9.16 & 241.67 & 78.64 & 6.97 \\
\midrule
\multicolumn{9}{l}{\textbf{\textit{The evolution of VUM model}}} \\
\textit{Config1}$^\dag$ & 258.7M & - & - & DiT-24/768 & 75$\%$ Mask+Noise & 316.9 & 2.07 & 74.41 & 6.57 \\
\textit{Config2} & 217.3M & ViT-12/768 & - & DiT-12/768 & 75$\%$ Mask+Noise & 159.6 & 5.66 & 80.63 & 21.25 \\
\textit{Config3} & 217.3M & ViT-12/768 & - & DiT-12/768 & Balanced HSNR & 22.07 & 80.50 & 81.69 & 43.25 \\
VUM & 243.3M & ViT-12/768 & ViT-8/512 & DiT-12/768 & Balanced HSNR & 9.77 & 193.51 & 81.79 & 48.70 \\ 
\bottomrule
\end{tabular}
\end{adjustbox}
\end{small}
\end{center}
\vspace{-0.4cm}
\end{table*}

\begin{figure*}[htbp]
\begin{minipage}{\textwidth}
    \begin{minipage}{0.68\textwidth}
        \captionof{table}{
        Ablations on the information-balanced scheduling of Balanced HSNR. We compare various scheduling functions that dictate the trajectory of the SNR across different noise levels.
        } 
        \label{tab:ablations_schedule}
        \vspace{-0.3cm}
        \begin{center}
        \begin{small}
        \begin{adjustbox}{max width=\textwidth}
        \begin{tabular}{l|cc|cc}
        \toprule
        Scheduling & FID$\downarrow$ & IS$\uparrow$ & IN1K-FT$\uparrow$ & IN1K-LP$\uparrow$ \\
        \midrule
        Sqrt: $r(t) = r_{\text{h}} - t \cdot \sqrt{r_{\text{h}} - r_{\text{l}}}$ 
        & 10.07 & 193.43 & 81.65 &  48.49 \\
        Square: $r(t) = r_{\text{h}} - t \cdot (r_{\text{h}} - r_{\text{l}})^2$
        & 23.18 & 76.63 & 81.59 & 41.66 \\
        Cosine: $r(t) = r_{\text{l}} + \frac{r_{\text{h}} - r_{\text{l}}}{2} \cdot (1 + \cos(\pi t))$
        & 10.30 & 185.47 & 81.77 & 48.51 \\
        Linear: $r(t) = r_{\text{h}} - t \cdot (r_{\text{h}} - r_{\text{l}})$ 
        & 9.77 & 193.51 & 81.79 & 48.70\\
        \bottomrule
        \end{tabular}
        \end{adjustbox}
        \end{small}
        \end{center}
        \vskip -0.1in
    \end{minipage}
    \hfill
    \begin{minipage}{0.3\textwidth}
        \includegraphics[width=\linewidth]{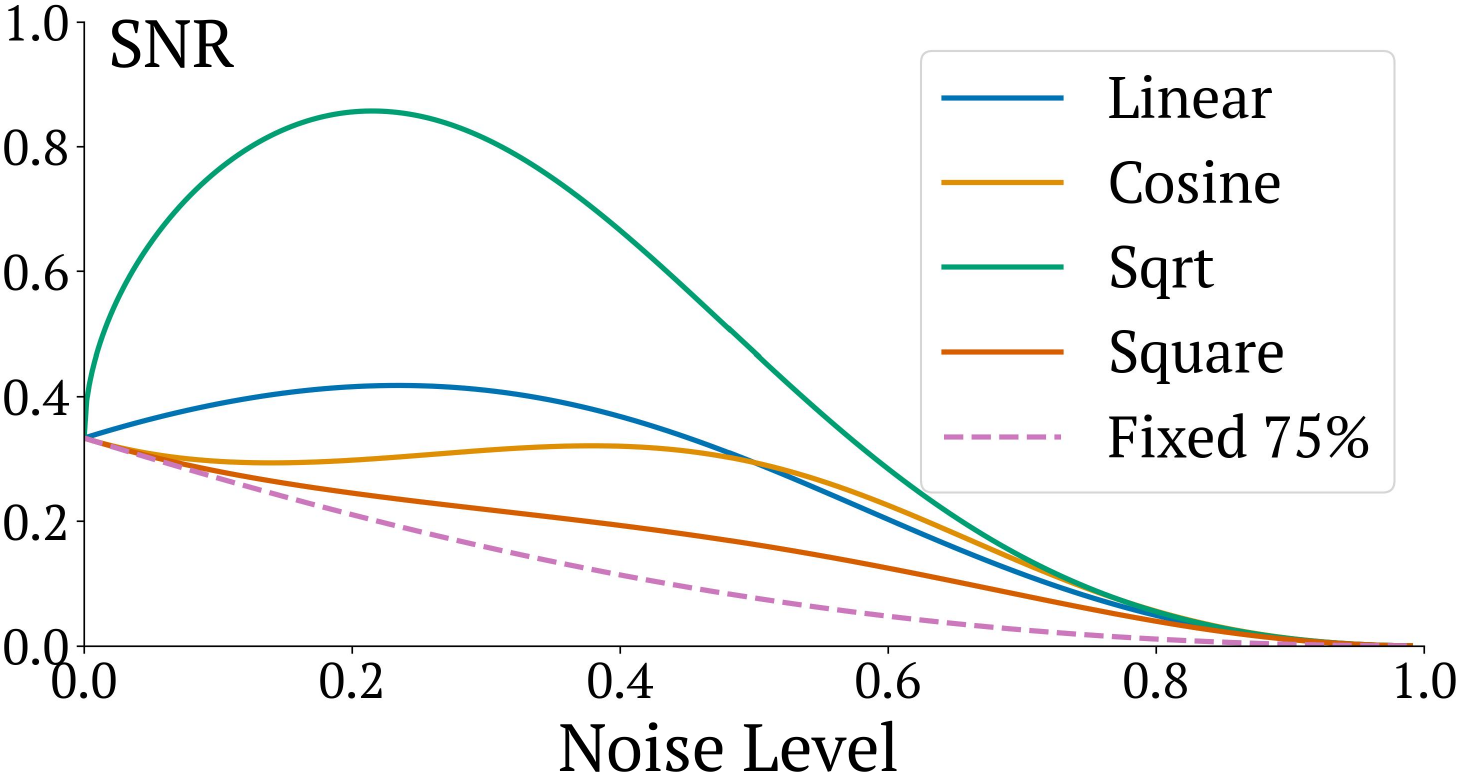}
        \caption{The SNR Curve of different scheduling functions.}
        \label{fig:snr_schedule}
        \vskip -0.2in
    \end{minipage}
\end{minipage}
\vspace{-0.2cm}
\end{figure*}

\begin{figure*}[htbp]
\begin{minipage}{\textwidth}
    \begin{minipage}{0.35\textwidth}
        \includegraphics[width=\linewidth]{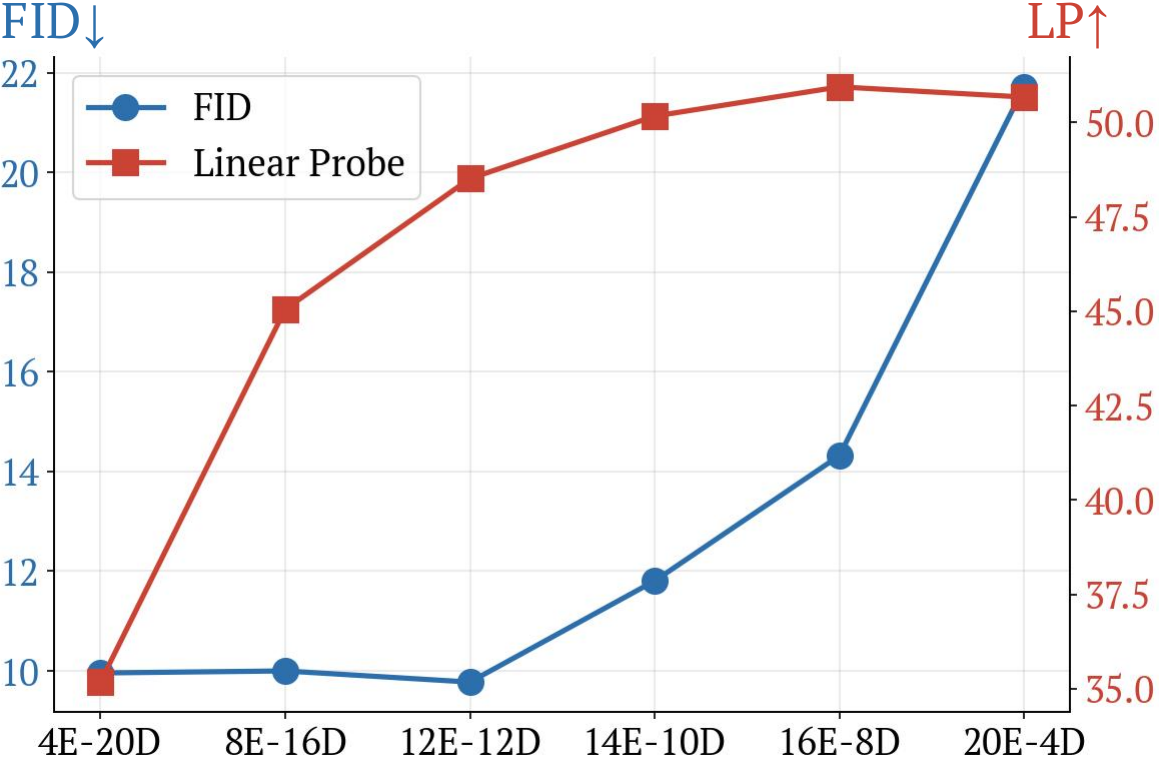}
        \caption{Ablations on the encoder-decoder layer ratio.} 
        \label{fig:ablations_depth}
        \vskip -0.2in
    \end{minipage}
    \hfill
    \begin{minipage}{0.3\textwidth}
        \captionof{table}{
        Ablations on the spatial decoupling weight $\lambda$ in \cref{eq:loss_generation}. The task-balancing weight is fixed at $\beta=1.0$.
        } 
        \label{tab:ablations_lambdda}
        \begin{center}
        \begin{small}
        \begin{adjustbox}{max width=\textwidth}
        \begin{tabular}{l|cc}
        \toprule
        $\lambda$ & FID$\downarrow$ & IN1K-LP$\uparrow$ \\
        \midrule
        0.0 & 9.52 & 22.79 \\
        0.1 & 9.77 & 48.70 \\
        0.5 & 12.42 & 46.36 \\
        1.0 & 13.41 & 48.08 \\
        \bottomrule
        \end{tabular}
        \end{adjustbox}
        \end{small}
        \end{center}
        \vskip -0.1in
    \end{minipage}
    \hfill
    \begin{minipage}{0.3\textwidth}
        \captionof{table}{
        Ablations on the task-balancing weight $\beta$ in \cref{eq:loss_total}. The spatial decoupling weight is fixed at the default $\lambda=0.1$.
        } 
        \label{tab:ablations_beta}
        \begin{center}
        \begin{small}
        \begin{adjustbox}{max width=\textwidth}
        \begin{tabular}{l|cc}
        \toprule
        $\beta$ & FID$\downarrow$ & IN1K-LP$\uparrow$ \\
        \midrule
        0.5 & 9.75 & 37.53 \\
        1.0 & 9.77 & 48.70 \\
        1.5 & 12.23 & 46.01 \\
        2.0 & 12.52 & 33.44 \\
        \bottomrule
        \end{tabular}
        \end{adjustbox}
        \end{small}
        \end{center}
        \vskip -0.1in
    \end{minipage}    
\end{minipage}
\vspace{-0.5cm}
\end{figure*}

\subsection{Ablation Studies}
\label{sec:ablation}

We conduct comprehensive ablations to validate the design choices of VUM. All model variants are trained on ImageNet-1K at $256 \times 256$ resolution for 160 epochs, and generative metrics are reported using a classifier-free guidance (CFG) scale of 2.5.
We ablate our framework by evaluating its architectural evolution from standard baselines (\cref{tab:ablations_component}), HSNR scheduling strategies (\cref{tab:ablations_schedule}), encoder-decoder depth ratios (\cref{fig:ablations_depth}), and loss-balancing coefficients ($\lambda, \beta$).

\noindent\textbf{Evolution of VUM.}
\cref{tab:ablations_component} traces our structural trajectory. Task-specific baselines expose inherent divergence: MAE excels at perception (IN1K-LP 44.85) but cannot generate, whereas JiT synthesizes well (FID 9.16) but suffers representational collapse (LP 6.97). Naive joint training (\textit{Config1}) fails catastrophically. Introducing an encoder-decoder (\textit{Config2}) only marginally recovers perception while struggling with synthesis (FID 159.6). Crucially, integrating the information-balanced HSNR (\textit{Config3}) bridges this gap, dramatically improving FID to 22.07 and LP to 43.25. Finally, the full VUM uses dual decoders to achieve a competitive balance: an FID of 9.77 compared with JiT's 9.16 and an IN1K-LP of 48.70 compared with the 160-epoch MAE baseline's 44.85, demonstrating generation and perception within one shared encoder.


\noindent\textbf{HSNR Scheduling Strategies.}
We compare four $r(t)$ decay functions (\cref{tab:ablations_schedule}, \cref{fig:snr_schedule}): our default Linear ($r_{\text{h}} - t(r_{\text{h}} - r_{\text{l}})$), Sqrt ($r_{\text{h}} - t\sqrt{r_{\text{h}} - r_{\text{l}}}$), Square ($r_{\text{h}} - t(r_{\text{h}} - r_{\text{l}})^2$), and Cosine ($r_{\text{l}} + \frac{r_{\text{h}} - r_{\text{l}}}{2}(1 + \cos(\pi t))$). The Square schedule drops SNR too rapidly, severely degrading both FID (23.18) and IN1K-LP (41.66). Conversely, the Sqrt schedule maintains an excessively high SNR, oversimplifying reconstruction, and hurting fidelity (FID 10.07). While Cosine transitions smoothly, the Linear schedule strictly dominates, yielding peak generative (FID 9.77) and semantic (LP 48.70) performance.


\noindent\textbf{Capacity Allocation.} Under a fixed 24-layer budget (\cref{fig:ablations_depth}), increasing encoder depth improves perception (LP), but overly compressing the decoder drastically degrades generation (FID). We find that a symmetric 12E-12D configuration yields the optimal trade-off between semantic abstraction and generative fidelity.


\noindent\textbf{Objective Balancing.} We ablate the spatial weight $\lambda$ (\cref{tab:ablations_lambdda}) and task weight $\beta$ (\cref{tab:ablations_beta}). Increasing $\lambda$ from 0 to 0.1 improves LP from 22.79 to 48.70, with a small FID increase from 9.52 to 9.77. Similarly, $\beta=1.0$ achieves the highest LP among the tested task weights while maintaining an FID of 9.77, illustrating the selected balance between generation and perception.

\section{Conclusion and Limitation}


We present the Visual Unified Model (VUM), a pixel-native model that jointly learns visual perception and image generation through a shared encoder. By conceptualizing generative noise and perceptual masking within a unified degradation continuum, VUM bridges both paradigms via an Information-Balanced (HSNR) schedule, task-adaptive routing, and an asymmetric dual-decoder topology. Without relying on external VAEs, VUM achieves competitive pixel-space generation, with ImageNet-256 FID close to JiT and DeCo (1.91 vs. 1.86 and 1.90), while the shared encoder transfers competitively to recognition and dense prediction. The ablations further show that coupling masking with the diffusion timestep and using an auxiliary perception decoder help balance generative fidelity and representation quality. These findings support further exploration of unified visual learning from raw pixels under a shared encoder.

\noindent\textbf{Limitations and Future Work.}
VUM's linear-probing gap relative to MAE motivates further study of noise-conditioned normalization and constraints on clean-state features. Scaling to larger models and higher resolutions also requires addressing the computational costs of pixel-space modeling. Our generation experiments are limited to class-conditional ImageNet; extending VUM to text-conditioned, web-scale multimodal learning remains future work.
On the perception side, we will also extend evaluation to depth and surface normal estimation.

\begin{ack}
This work was supported by the JC STEM Lab of AI for Science and Engineering, funded by The Hong Kong Jockey Club Charities Trust and the Research Grants Council of Hong Kong (Project No. CUHK14213224). No competing interests are declared for this work.
\end{ack}

{\small
\bibliographystyle{plain}
\bibliography{main}
}

\clearpage
\appendix
\section*{Appendix}
We include additional derivations, experimental details, and results in the appendix. 
\begin{itemize}
    \item In \cref{app:snr_derivation}, we provide a detailed derivation of Hybrid Signal-to-Noise Ratio (HSNR).
    \item In \cref{app:extened_related_work}, we we discuss more related works, including generative models with external representation and masked image modeling for image generation.
    \item In \cref{app:more_implementation_details}, we provide more implementation details of experiments, including model architecture, training procedure, and downstream perception evaluations. 
    \item In \cref{app:additional_results}, we provide additional empirical results, including HSNR comparisons, the gradient conflict analysis and structure target analysis.
    \item In \cref{app:feature_analysis}, we compare the feature map of VUM with MAE and JiT.
    \item In \cref{app:qualitative_results}, we provide more qualitative results of VUM. 
\end{itemize}


\section{Mathematical Derivation of Hybrid SNR}
\label{app:snr_derivation}

In this section, we provide the rigorous mathematical derivations for the independent Signal-to-Noise Ratios (SNR) of continuous diffusion and discrete masking, and formally deduce our proposed Hybrid SNR (HSNR) formulation.

\subsection{Preliminary Definitions}
Let the original clean image be a random variable $\mathbf{x} \sim p_{\text{data}}(\mathbf{x})$ drawn from the natural image distribution. For simplicity, we assume the data is pre-normalized to have zero mean and unit variance, such that the expected signal power is $\mathbb{E}[\mathbf{x}^2] = 1$. Let the continuous degradation be driven by Gaussian noise $\boldsymbol{\epsilon} \sim \mathcal{N}(\mathbf{0}, \mathbf{I})$ with $\mathbb{E}[\boldsymbol{\epsilon}^2] = 1$. Let the spatial masking be defined by an independent Bernoulli random variable $\mathbf{m} \in \{0, 1\}$, where the probability of a patch being masked is $r$, and the probability of remaining visible is $1-r$. Thus, the expectation of the mask indicator is $\mathbb{E}[\mathbf{m}] = 1-r$, and the expectation of the masked region indicator is $\mathbb{E}[1-\mathbf{m}] = r$.

\subsection{Independent SNRs}
\paragraph{Continuous Diffusion SNR.} 
For a standard continuous noise schedule, the degraded state at trajectory step $t \in [0, 1]$ is defined as:
\begin{equation}
    \mathbf{x}_t = (1-t)\mathbf{x} + t\boldsymbol{\epsilon}
\end{equation}
The preserved signal power $P_{\text{sig}}^{\text{diff}}$ and the additive noise power $P_{\text{noise}}^{\text{diff}}$ are calculated as:
\begin{equation}
    P_{\text{sig}}^{\text{diff}} = \mathbb{E}[((1-t)\mathbf{x})^2] = (1-t)^2, \quad P_{\text{noise}}^{\text{diff}} = \mathbb{E}[(t\boldsymbol{\epsilon})^2] = t^2
\end{equation}
Thus, the diffusion SNR is strictly formulated as:
\begin{equation}
    SNR_{\text{Diff}}(t) = \frac{P_{\text{sig}}^{\text{diff}}}{P_{\text{noise}}^{\text{diff}}} = \frac{(1-t)^2}{t^2}
\end{equation}

\paragraph{Discrete Masking SNR.} 
For pure spatial masking, the observed state is $\mathbf{x}_{\text{mask}} = \mathbf{m} \odot \mathbf{x}$. The SNR is defined by the ratio of the visible signal power to the spatial information loss (which acts as the reconstruction uncertainty or ``noise'' from the perspective of the decoder):
\begin{align}
    P_{\text{sig}}^{\text{mask}} &= \mathbb{E}[\mathbf{m} \mathbf{x}^2] = \mathbb{E}[\mathbf{m}]\mathbb{E}[\mathbf{x}^2] = 1-r \\
    P_{\text{loss}}^{\text{mask}} &= \mathbb{E}[(1-\mathbf{m})\mathbf{x}^2] = \mathbb{E}[1-\mathbf{m}]\mathbb{E}[\mathbf{x}^2] = r
\end{align}
Consequently, the masking SNR is:
\begin{equation}
    SNR_{\text{MAE}}(r) = \frac{P_{\text{sig}}^{\text{mask}}}{P_{\text{loss}}^{\text{mask}}} = \frac{1-r}{r}
\end{equation}

\subsection{Hybrid SNR (Joint Degradation)}
To model the unified continuum, we superimpose the spatial mask $r$ onto the noisy continuous manifold $\mathbf{x}_t$. The joint degraded state is formulated as:
\begin{equation}
    \mathbf{x}_{r,t} = \mathbf{m} \odot \mathbf{x}_t = \mathbf{m} \odot ((1-t)\mathbf{x} + t\boldsymbol{\epsilon})
\end{equation}
To derive the Hybrid SNR, we must calculate the joint preserved signal power and the aggregate uncertainty. 

\textbf{Joint Preserved Signal Power.} The original signal $(1-t)\mathbf{x}$ is only preserved in the visible patches ($\mathbf{m}=1$). Since the mask $\mathbf{m}$ and the signal $\mathbf{x}$ are independent, the expected preserved signal power is:
\begin{equation}
    P_{\text{sig}}^{\text{hybrid}} = \mathbb{E}\left[\mathbf{m} ((1-t)\mathbf{x})^2\right] = \mathbb{E}[\mathbf{m}] \cdot (1-t)^2 \cdot \mathbb{E}[\mathbf{x}^2] = (1-r)(1-t)^2
\end{equation}

\textbf{Aggregate Uncertainty.} The total uncertainty to the reconstruction objective stems from two mutually exclusive sources: the additive Gaussian noise present in the visible regions, and the complete structural signal loss in the masked regions.
\begin{itemize}
    \item \textit{Visible Continuous Noise:} The variance of the Gaussian noise within the unmasked patches is:
    \begin{equation}
        P_{\text{noise}}^{\text{visible}} = \mathbb{E}\left[\mathbf{m} (t\boldsymbol{\epsilon})^2\right] = \mathbb{E}[\mathbf{m}] \cdot t^2 \cdot \mathbb{E}[\boldsymbol{\epsilon}^2] = (1-r)t^2
    \end{equation}
    \item \textit{Discrete Spatial Loss:} The missing signal in the masked patches acts as absolute uncertainty. The variance of this completely lost signal is:
    \begin{equation}
        P_{\text{loss}}^{\text{spatial}} = \mathbb{E}\left[(1-\mathbf{m})\mathbf{x}^2\right] = \mathbb{E}[1-\mathbf{m}] \cdot \mathbb{E}[\mathbf{x}^2] = r
    \end{equation}
\end{itemize}
Because spatial masking creates disjoint sets of pixels, these two uncertainties aggregate linearly to form the total reconstruction noise power:
\begin{equation}
    P_{\text{uncertainty}}^{\text{total}} = P_{\text{noise}}^{\text{visible}} + P_{\text{loss}}^{\text{spatial}} = (1-r)t^2 + r
\end{equation}

\textbf{Final Formulation.} By dividing the joint preserved signal power by the aggregate uncertainty, we obtain the mathematically exact Hybrid SNR:
\begin{equation}
    SNR_{\text{Hybrid}}(r, t) = \frac{P_{\text{sig}}^{\text{hybrid}}}{P_{\text{uncertainty}}^{\text{total}}} = \frac{(1-r)(1-t)^2}{(1-r)t^2 + r}
    \label{eq:snr_hybrid_appendix}
\end{equation}
This formulation smoothly degenerates to $SNR_{\text{Diff}}(t)$ when $r=0$, and to $SNR_{\text{MAE}}(r)$ when $t=0$, mathematically proving that spatial masking and Gaussian noising exist on the exact same continuous degradation manifold.

\section{Extended Related Work}
\label{app:extened_related_work}

\paragraph{Generative Models with External Representations.} 


Recent advancements have increasingly leveraged features from pre-trained Vision Foundation Models (VFMs) to augment diffusion training. For example, REPA~\cite{yu2024repa} enhances semantic representation quality by enforcing the alignment of features between the early diffusion layers and the features of frozen vision. Based on this, REPA-E~\cite{leng2025repa-e} extends the framework to support end-to-end VAE fine-tuning, further improving the alignment of features with VFM. REG~\cite{wu2025reg} integrates spatial visual features with semantic class embeddings derived from VFMs, establishing a persistent semantic steering mechanism throughout the generative process.
At the tokenizer level, ReaLS~\cite{xu2025reals} and the VA-VAE used by LightningDiT~\cite{yao2025lightningdit} distill semantic priors from pre-trained VFMs into VAE encoders through representation alignment. Diffusion transformers are subsequently trained in the latent spaces produced by these aligned encoders.
Despite their efficacy, these methodologies share a fundamental bottleneck: they strictly depend on external, off-the-shelf vision encoders to supply semantic priors. In contrast, our framework entirely discards the reliance on external representations; instead, by jointly optimizing for signal recovery and generation, we compel the model to intrinsically develop powerful semantic abstractions directly from raw visual signals.

\paragraph{Masked Image Modeling for Image Generation.} 
Masked image modeling (MIM)~\cite{bao2021beit,Xie2021simmim,He2021mae} has been increasingly adapted for image synthesis. Early approaches operate in discrete token spaces; for instance, MaskGIT~\cite{chang2022maskgit} employs a bidirectional transformer with an iterative parallel decoding strategy, while MAGE~\cite{li2023mage} utilizes variable masking ratios to train both generation and representation capabilities within a single framework. Moving beyond discrete tokenization, MAR~\cite{li2024mar} performs masked autoregressive modeling in continuous-valued spaces by integrating a diffusion loss. FractalMAR~\cite{li2025fractalmar} further extends this to the pixel level through a recursive, divide-and-conquer fractal architecture. An orthogonal line of work integrates masking mechanisms into diffusion models. MaskDiT~\cite{zheng2023maskdit} applies an asymmetric encoder-decoder architecture to process only unmasked patches during diffusion training, and MDT~\cite{gao2023mdt} incorporates a mask latent modeling scheme to predict masked tokens from unmasked ones within the diffusion process. Unlike these approaches that either rely on discrete tokenizers or treat masking merely as an auxiliary mechanism, VUM fundamentally integrates masked image modeling and continuous diffusion into a joint signal recovery objective, achieving true perception-generation unification directly in the pure pixel space.

\section{More Implementation Details}
\label{app:more_implementation_details}

\begin{table}[ht]
\caption{\textbf{Architectural and optimization details for VUM-B and VUM-L.} The framework employs a shared encoder paired with asymmetric decoupled decoders.}
\label{tab:vum_details}
\centering
\begin{tabular}{lcc}
\toprule
& \textbf{VUM-B} & \textbf{VUM-L} \\
\midrule
\textbf{Shared ViT Encoder ($\mathcal{E}$)} & & \\
Input resolution & $256 \times 256 \times 3$ & $256 \times 256 \times 3$ \\
Patch size & $16 \times 16$ & $16 \times 16$ \\
Num. layers & 12 & 24 \\
Hidden dim. & 768 & 1024 \\
Num. heads & 12 & 16 \\
Parameters & 85.2M & 302.5M \\
\midrule
\textbf{Heavy DiT Decoder ($\mathcal{D}_\text{G}$)} & & \\
Num. layers & 12 & 24 \\
Hidden dim. & 768 & 1024 \\
Num. heads & 12 & 16 \\
Parameters & 130.5M & 458.1M \\
\midrule
\textbf{Auxiliary ViT Decoder ($\mathcal{D}_\text{P}$)} & & \\
Num. layers & 8 & 8 \\
Hidden dim. & 512 & 512 \\
Num. heads & 16 & 16 \\
Parameters & \multicolumn{2}{c}{26.1M} \\
\midrule
\textbf{Optimization} & & \\
Optimizer & \multicolumn{2}{c}{Muon} \\
Learning rate & \multicolumn{2}{c}{0.0002} \\
Learning rate Schedule & \multicolumn{2}{c}{Constant with warmup} \\
Weight decay & \multicolumn{2}{c}{0.01} \\
Batch size & \multicolumn{2}{c}{2048} \\
Warmup epochs & \multicolumn{2}{c}{40} \\
Training epochs & 160 & 800 \\
\bottomrule
\end{tabular}
\end{table}

\paragraph{Model Architecture.} We adopt the standard Vision Transformer (ViT) architecture for our shared encoder and auxiliary perceptual decoder, and the Diffusion Transformer (DiT) architecture for our heavy generative decoder. To incorporate conditional information during the generative phase, the DiT decoder utilizes adaptive layer normalization (adaLN) to condition the class embeddings. Instead of standard absolute positional embeddings, we apply 2D Rotary Position Embedding (2D RoPE)~\cite{su2024rope} across the network to better capture relative spatial relationships and enhance structural generalization.

\paragraph{Training.} For training, we utilize the Muon optimizer. Data augmentation is kept strictly minimal; in contrast to the \textit{RandomResizedCrop} typically employed by MAE, we uniformly apply a standard center-crop strategy alongside random horizontal flipping to process all input images. Crucially, to bridge the distribution gap with standard MAE inference, which operates on clean pixels without noise, we explicitly enforce that $25\%$ of the images in every training batch are kept perfectly clean ($t=0$) and subjected to a heavy $75\%$ spatial mask. This guaranties that a consistent portion of the gradient updates is dedicated entirely to the perceptual abstraction objective.

\paragraph{Generation Sampling.} For image synthesis, we employ an ODE-based sampler. We configure the generation process with 50 sampling steps and apply Classifier-Free Guidance (CFG) to enhance sample fidelity and class alignment. As CFG requires evaluating both the conditional and unconditional states at each step, this setup results in a total of 100 Number of Function Evaluations (NFE).

\paragraph{Linear Probing.} 
Following the established protocols in MAE~\cite{He2021mae}, we train a linear classifier on the frozen representations extracted from our shared ViT encoder. We optimize the linear head using the AdamW optimizer with a base learning rate of 0.1, a batch size of 16384, and a cosine decay schedule over 90 epochs (including 10 warmup epochs). We explicitly disable mixup, cutmix, drop path, and color jittering, and strictly set the weight decay to zero.

\paragraph{End-to-End Finetuning.} 
For full fine-tuning, we update all model parameters following MAE practices~\cite{He2021mae}. We employ the AdamW optimizer with a base learning rate of 1e-3, a weight decay of 0.05, and a global batch size of 1024. The training spans 100 epochs for VUM-B and 50 epochs for VUM-L, utilizing a cosine learning rate schedule with 5 warmup epochs. To effectively adapt the pre-trained representations, we apply a layer-wise learning rate decay of 0.75 across the shared encoder. Standard augmentations and regularizations are adopted, including RandAug (9, 0.5), label smoothing (0.1), mixup (0.8), cutmix (1.0), and drop path (0.1).

\paragraph{Object Detection and Instance Segmentation on COCO.} 
Following the adaptation strategy in MAE~\cite{He2021mae}, we integrate our VUM ViT encoder as the backbone for Mask R-CNN. Strictly following the implementation in \cite{chen2024cae}, we equally divide the Transformer blocks into four subsets and apply intermediate convolutions to extract multi-scale feature maps (strides 4, 8, 16, and 32) for the Feature Pyramid Network (FPN), fine-tuning the entire model for 12 epochs with a learning rate of $1.5\times10^{-4}$.

\paragraph{Semantic Segmentation on ADE20K.} 
We evaluate dense semantic segmentation performance using the UperNet framework. Following the implementation in \cite{chen2024cae}, the VUM ViT encoder serves as the backbone, and the entire network is fine-tuned end-to-end for 100 epochs with a batch size of 16 and a learning rate of $3.5\times10^{-4}$.

\section{Additional Results}
\label{app:additional_results}

\subsection{HSNR versus Loss Balancing and Training Curricula}
\label{app:hsnr_comparison}

To examine whether adaptive loss weighting or generic training curricula can replace HSNR, we compare five strategies under a matched architecture and training budget. All strategies use a ViT-B encoder and a DiT-B decoder, without the auxiliary ViT decoder, and are trained on ImageNet for 160 epochs. The static baseline and HSNR correspond to \textit{Config2} and \textit{Config3} in \cref{tab:ablations_component}, respectively.

\begin{itemize}
    \item \textbf{Static mask and noise.} At each training step, we apply a fixed $75\%$ random mask together with randomly sampled diffusion noise. This is the joint recovery baseline without information-balanced coupling.
    \item \textbf{Uncertainty weighting.} We retain the same static degradation as \textit{Config2}, but replace the fixed task weighting in \cref{eq:loss_total} with homoscedastic uncertainty weighting~\cite{kendall2018uncertainty}. Specifically, we learn two scalars $s_{\mathrm{G}}=\log\sigma_{\mathrm{G}}^2$ and $s_{\mathrm{P}}=\log\sigma_{\mathrm{P}}^2$ and optimize
    \begin{equation}
        \gL_{\mathrm{UW}} = \frac{1}{2}e^{-s_{\mathrm{G}}}\gL_{\mathrm{G}} + \frac{1}{2}e^{-s_{\mathrm{P}}}\gL_{\mathrm{P}} + \frac{1}{2}s_{\mathrm{G}} + \frac{1}{2}s_{\mathrm{P}}.
        \label{eq:uncertainty_weighting}
    \end{equation}
    The exponential terms adaptively weight the generative and perceptual losses, while the log-variance terms regularize these weights. This tests whether adaptive loss balancing alone can improve the static joint objective.
    \item \textbf{Progressive masking.} The diffusion noise level $t$ is sampled randomly, while the mask ratio is linearly annealed over training epochs from $0.75$ to $0.25$. This curriculum changes masking over the course of training without coupling it to the instantaneous noise level.
    \item \textbf{Stage-wise training.} We divide the same total epoch budget into two equal halves. The first half trains perception only, using a fixed $75\%$ mask and no diffusion noise. The second half trains diffusion only, using randomly sampled noise and no masking. This tests a sequential perception-to-generation curriculum.
    \item \textbf{HSNR.} We couple the mask ratio to the instantaneous noise level using \cref{eq:dynamic_r}, guided by the Hybrid SNR formulation in \cref{eq:snr_hybrid}. Unlike an epoch-based curriculum, this strategy adjusts spatial masking to the continuous corruption level throughout joint training.
\end{itemize}

\begin{table}[t]
\centering
\caption{\textbf{HSNR versus loss balancing and training curricula.} All strategies use a ViT-B encoder and a DiT-B decoder without an auxiliary ViT decoder, trained on ImageNet for 160 epochs. The static baseline and HSNR rows reuse \textit{Config2} and \textit{Config3} from \cref{tab:ablations_component}.}
\label{tab:hsnr_comparison}
\begin{small}
\begin{tabular}{lcc}
\toprule
Strategy & FID$\downarrow$ & IN1K-LP$\uparrow$ \\
\midrule
Static $75\%$ mask + noise & 159.6 & 21.25 \\
Uncertainty weighting & 150.35 & 21.93 \\
Progressive masking & 52.47 & 23.37 \\
Stage-wise (perception $\to$ generation) & 27.36 & 12.63 \\
\midrule
HSNR (\textit{Config3}) & \textbf{22.07} & \textbf{43.25} \\
\bottomrule
\end{tabular}
\end{small}
\end{table}

\paragraph{Analysis.}
As shown in \cref{tab:hsnr_comparison}, uncertainty weighting provides limited improvements over the static baseline: FID changes from $159.6$ to $150.35$, and IN1K-LP from $21.25$ to $21.93$. Adaptive loss weighting therefore does not recover the performance of HSNR under this setting. Progressive masking improves FID to $52.47$, but its LP remains $23.37$. Stage-wise training achieves a lower FID of $27.36$, while its LP falls to $12.63$, indicating a weaker final perceptual representation. Among these five strategies, HSNR achieves both the lowest FID ($22.07$) and the highest LP ($43.25$). These results support coupling input corruption to the current noise level, beyond reweighting task losses or scheduling them across epochs. With the auxiliary ViT decoder added, the full VUM further achieves FID $9.77$ and IN1K-LP $48.70$, as reported in \cref{tab:ablations_component}.

\subsection{Gradient Conflict Analysis}
\label{app:gradient_conflict}

To quantify the optimization interference between the $\mathcal{L}_\text{P}$ and $\mathcal{L}_\text{G}$ objectives in \cref{eq:loss_total}, we measure gradient conflict in the shared parameters. Given a mini-batch, we compute the MAE loss
$\mathcal{L}_{\text{P}}$ and diffusion loss $\mathcal{L}_{\text{G}}$ separately, obtaining their respective gradients $\mathbf{g}_{\text{P}} = \nabla\theta
\mathcal{L}_{\text{P}}$ and $\mathbf{g}_{\text{G}} = \nabla_\theta \mathcal{L}_{\text{G}}$ for each parameter group $\theta$. We adopt two complementary metrics to evaluate the conflicts. 

\paragraph{Gradient Cosine Similarity.} We measure the cosine similarity between the two gradient vectors:
\begin{equation}
    \text{CosSim}(\mathbf{g}_{\text{P}}, \mathbf{g}_{\text{G}}) = \frac{\mathbf{g}_{\text{P}} \cdot \mathbf{g}_{\text{G}}}{|\mathbf{g}_{\text{P}}| \cdot |\mathbf{g}_{\text{G}}|}
\end{equation}
which measures the overall directional alignment between the two gradient vectors. A positive value indicates cooperative optimization, zero indicates orthogonality, and a negative
value indicates conflicting objectives.

\paragraph{Sign Conflict Ratio.} The sign conflict ratio is defined as the fraction of parameter elements (among those with magnitude above the median threshold for both gradients) that receive opposing gradient directions from the two tasks.  To suppress noise from near-zero gradient elements, we define the significant set $\mathcal{S} = \{i : |g_{\text{P}}^{(i)}| >
\text{med}(|\mathbf{g}_{\text{P}}|) \wedge |\mathbf{g}_{\text{G}}^{(i)}| > \text{med}(|\mathbf{g}_{\text{diff}}|)\}$, where $\text{med}(\cdot)$ denotes the element-wise median. The sign
conflict ratio is then:
\begin{equation}
    \text{SCR} = \frac{\big|\{i \in \mathcal{S} : \text{sign}(g_{\text{P}}^{(i)}) \neq \text{sign}(g_{\text{G}}^{(i)})\}\big|}{|\mathcal{S}|}
\end{equation}
This metric captures the fraction of significant parameters that receive opposing update directions from the two tasks.

\begin{figure*}[htbp]
\begin{center}
\centerline{\includegraphics[width=1.0\linewidth]{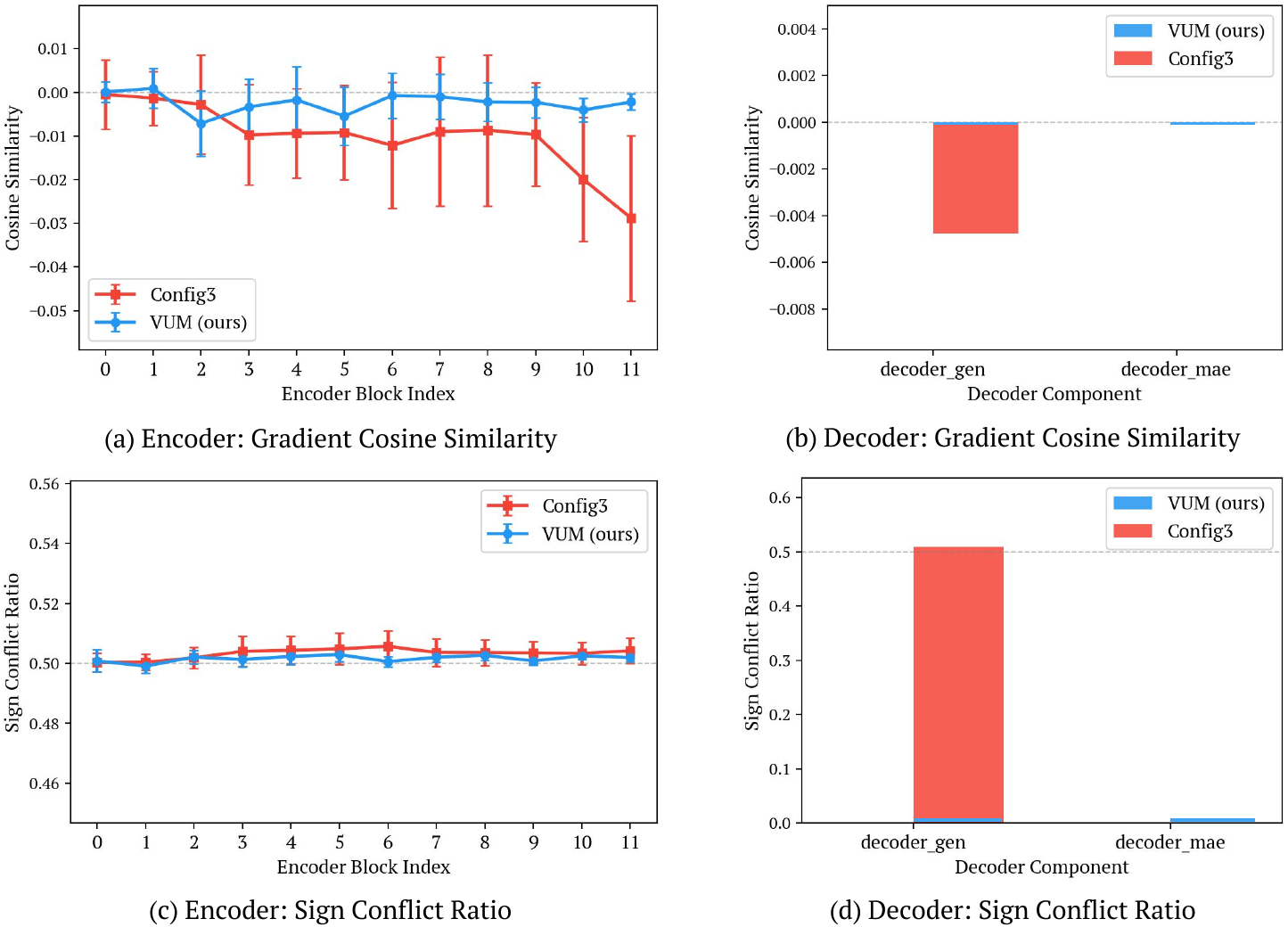}}
\caption{\textbf{Gradient conflict analysis between perception and generation objectives.} We compare our decoupled VUM architecture against a single-decoder baseline (\textit{Config 3} in \cref{tab:ablations_component}) across the shared encoder layers (a, c) and the decoder components (b, d). By physically decoupling the decoding phase, VUM perfectly isolates task-specific gradients in the decoders, thereby shielding the shared encoder and maintaining conflict-free optimization.}
\label{fig:gradient_conflict}
\end{center}
\end{figure*}

\paragraph{Analysis.} We analyze our VUM-B with its single-decoder counterpart (i.e., \textit{Config 3} in \cref{tab:ablations_component}). As shown in \cref{fig:gradient_conflict} (a) and (c), the single-decoder baseline suffers from severe optimization interference within the shared encoder. Specifically, its cosine similarity progressively drops into the negative regime, and the SCR consistently exceeds $0.5$, with the conflict peaking in the deeper encoder layers (blocks $8$--$11$). This demonstrates that the perception and generation objectives actively pull the shared representations in opposing directions. 

In stark contrast, VUM successfully neutralizes this interference, maintaining near-perfect gradient orthogonality (cosine similarity $\approx 0$ and SCR $\approx 0.5$) across all encoder blocks. The decoder-level analysis (b, d) exposes the mechanical root of this success: while the unified decoder in \textit{Config 3} faces catastrophic gradient collisions (massive negative cosine similarity and high SCR), VUM's decoupled dual-decoders physically isolate the backward passes. Because $\mathcal{L}_{\text{P}}$ and $\mathcal{L}_{\text{G}}$ only route through their dedicated decoders ($\gD_{\text{P}}$ and $\gD_{\text{G}}$), the parameter-level gradient overlap is strictly zero. This confirms our architectural motivation: the decoupled decoders function as highly effective structural firewalls, completely eradicating task-level gradient conflicts before they can pollute the shared encoder.

\subsection{Extra Ablations}

Across the stochastic continuum ($t>0$), a heavy DiT drives the spatial synthesis. To handle variable masking at arbitrary noise levels, $\gD_{\text{G}}$ is governed by a state-conditioned regional objective (\cref{eq:loss_generation}): it applies standard denoising to the visible regions while actively reconstructing the \textit{noisy} structure for the masked regions. Crucially, by targeting the noisy rather than the clean structure, the decoder natively preserves spatial coherence within the current temporal state $t$. The necessity of this trajectory-aligned target is explicitly ablated in \cref{tab:ablations_mask_region}.

Compared to the default loss \cref{eq:loss_generation}, its clean structure variant is defined as follows:
\begin{equation}
    \gL_{\text{G}} = \E_{t} \left[ \| (\x - \gD_{\text{G}}([\vh, \vs], t)) \odot (1-\mM) \|_2^2 + \lambda \| (\texttt{Norm}(\x) - \gD_{\text{G}}([\vh, \vs], t)) \odot \mM \|_2^2 \right],
    \label{eq:loss_generation_clean}
\end{equation}

\paragraph{Analysis} As detailed in \cref{tab:ablations_mask_region}, the choice of reconstruction target for the masked regions during the generative phase ($t>0$) profoundly impacts the network's capabilities. We observe a severe degradation in generative performance when $\gD_{\text{G}}$ is forced to predict the clean structure, with FID surging from $9.77$ to $16.18$ and IS plummeting from $193.51$ to $131.54$. Furthermore, the perceptual metrics (IN1K-FT and IN1K-LP) also suffer noticeable declines.

This stark performance gap is rooted in the physical consistency of the degradation trajectory. When targeting the clean structure (\cref{eq:loss_generation_clean}), the network is compelled to execute two mathematically conflicting operations within the exact same feature map: standard incremental denoising for visible patches at timestep $t$, and a premature, full-trajectory projection to $t=0$ for the masked patches. This severe spatial inconsistency essentially fractures the unified degradation continuum.

Conversely, targeting the noisy structure (\cref{eq:loss_generation}) guarantees that the reconstructed masked latents remain safely anchored to the exact same stochastic trajectory as the visible patches. By maintaining this strict spatial and temporal coherence at any given state $t$, VUM profoundly stabilizes the generative optimization. This trajectory-aligned strategy is the key to unlocking the model's full potential, delivering a substantial boost in both high-fidelity synthesis and discriminative abstraction.

\begin{table*}[t]
\caption{\textbf{Ablation on the masked region reconstruction target during the generative phase ($t>0$).} Forcing the DiT decoder to predict the clean structure (Eq. \ref{eq:loss_generation_clean}) severely degrades generation fidelity and representation quality. In contrast, targeting the noisy structure (Eq. \ref{eq:loss_generation}) intrinsically aligns the masked latents with the current stochastic trajectory, yielding drastically superior performance across all metrics.}
\label{tab:ablations_mask_region}
\begin{center}
\begin{small}
\begin{adjustbox}{max width=\textwidth}
\begin{tabular}{l|cc|cc}
\toprule
Mask Region Target & FID$\downarrow$ & IS$\uparrow$ & IN1K-FT$\uparrow$ & IN1K-LP$\uparrow$ \\
\midrule
Clean Structure~\cref{eq:loss_generation_clean} & 16.18 & 131.54 & 80.57 & 41.47 \\
Noisy Structure~\cref{eq:loss_generation} & 9.77 & 193.51 & 81.79 & 48.70\\
\bottomrule
\end{tabular}
\end{adjustbox}
\end{small}
\end{center}
\end{table*}

\section{Feature Analysis}
\label{app:feature_analysis}

\begin{figure*}[htbp]
\begin{center}
\centerline{\includegraphics[width=1.0\linewidth]{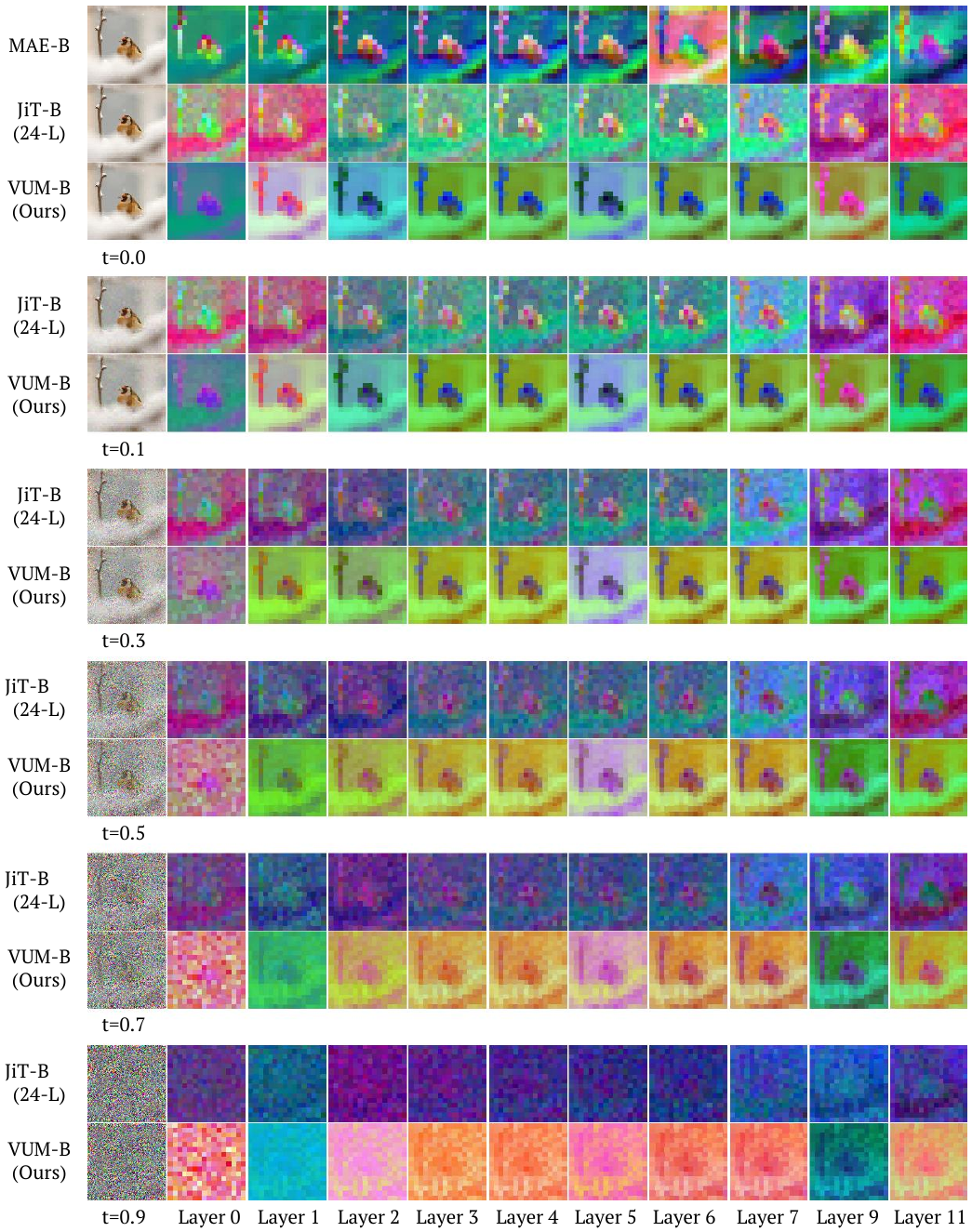}}
\caption{PCA feature map comparison among MAE-B, JiT-B (24 layers), and VUM-B. MAE-B is evaluated exclusively on clean images ($t=0$), whereas JiT-B and VUM-B are evaluated across varying noise levels $t \in \{0, 0.1, 0.3, 0.5, 0.7, 0.9\}$. For all models, features are extracted from layers $\{0, 1, 2, 3, 4, 5, 6, 7, 9, 11\}$.}
\label{fig:pca_1}
\end{center}
\end{figure*}

\begin{figure*}[htbp]
\begin{center}
\centerline{\includegraphics[width=1.0\linewidth]{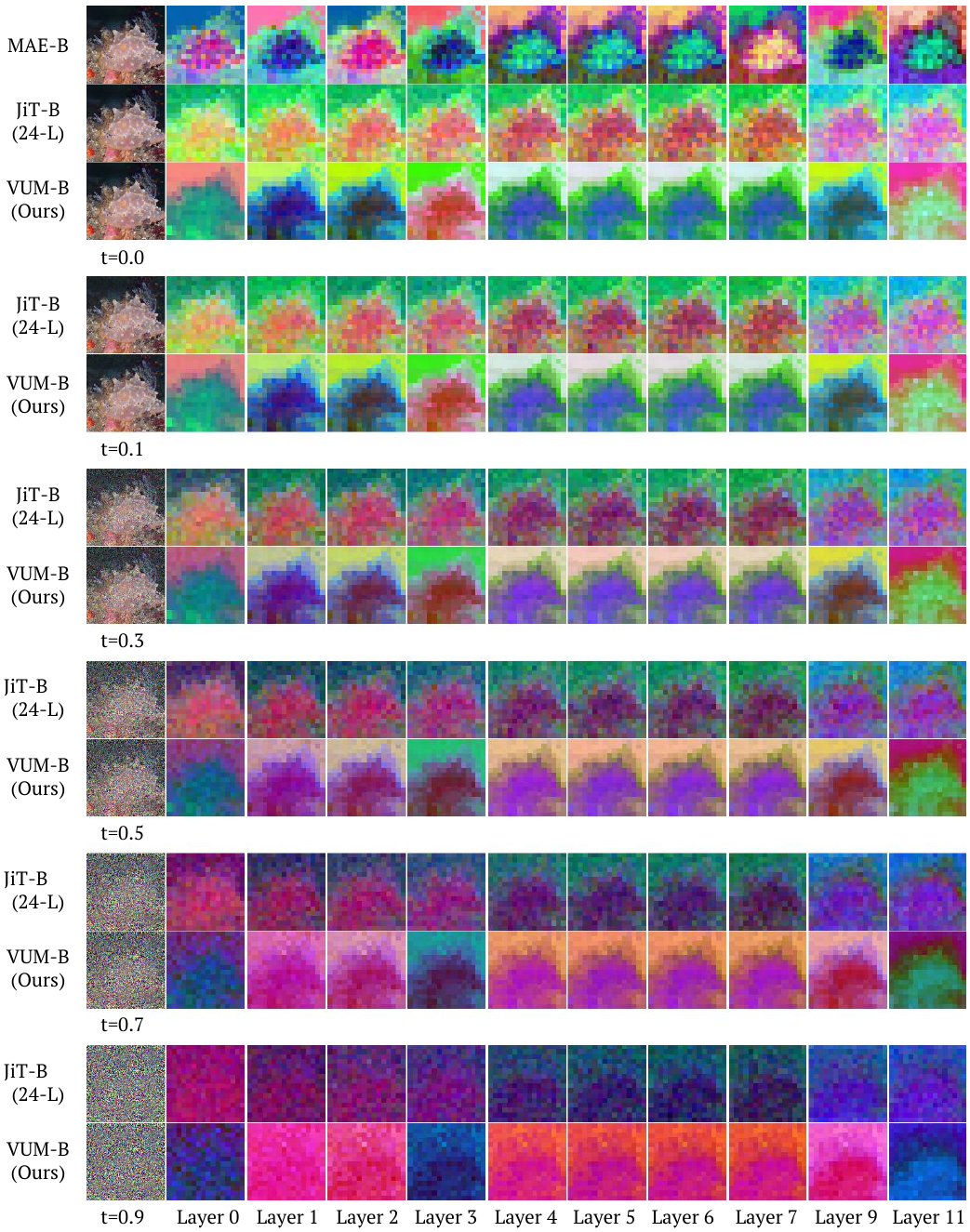}}
\caption{PCA feature map comparison among MAE-B, JiT-B (24 layers), and VUM-B. MAE-B is evaluated exclusively on clean images ($t=0$), whereas JiT-B and VUM-B are evaluated across varying noise levels $t \in \{0, 0.1, 0.3, 0.5, 0.7, 0.9\}$. For all models, features are extracted from layers $\{0, 1, 2, 3, 4, 5, 6, 7, 9, 11\}$.}
\label{fig:pca_2}
\end{center}
\end{figure*}

\begin{figure*}[htbp]
\begin{center}
\centerline{\includegraphics[width=1.0\linewidth]{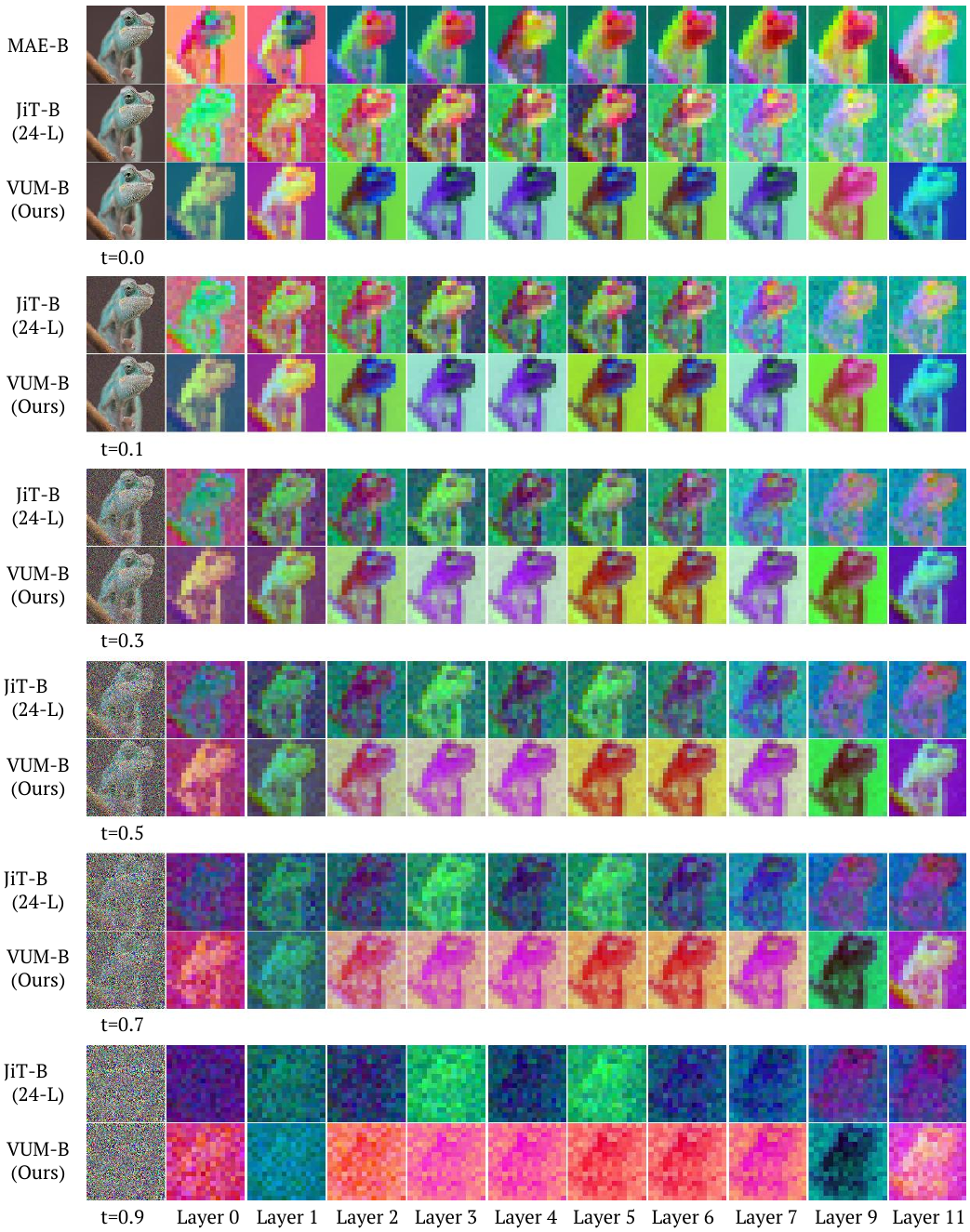}}
\caption{PCA feature map comparison among MAE-B, JiT-B (24 layers), and VUM-B. MAE-B is evaluated exclusively on clean images ($t=0$), whereas JiT-B and VUM-B are evaluated across varying noise levels $t \in \{0, 0.1, 0.3, 0.5, 0.7, 0.9\}$. For all models, features are extracted from layers $\{0, 1, 2, 3, 4, 5, 6, 7, 9, 11\}$.}
\label{fig:pca_3}
\end{center}
\end{figure*}

\begin{figure*}[htbp]
\begin{center}
\centerline{\includegraphics[width=1.0\linewidth]{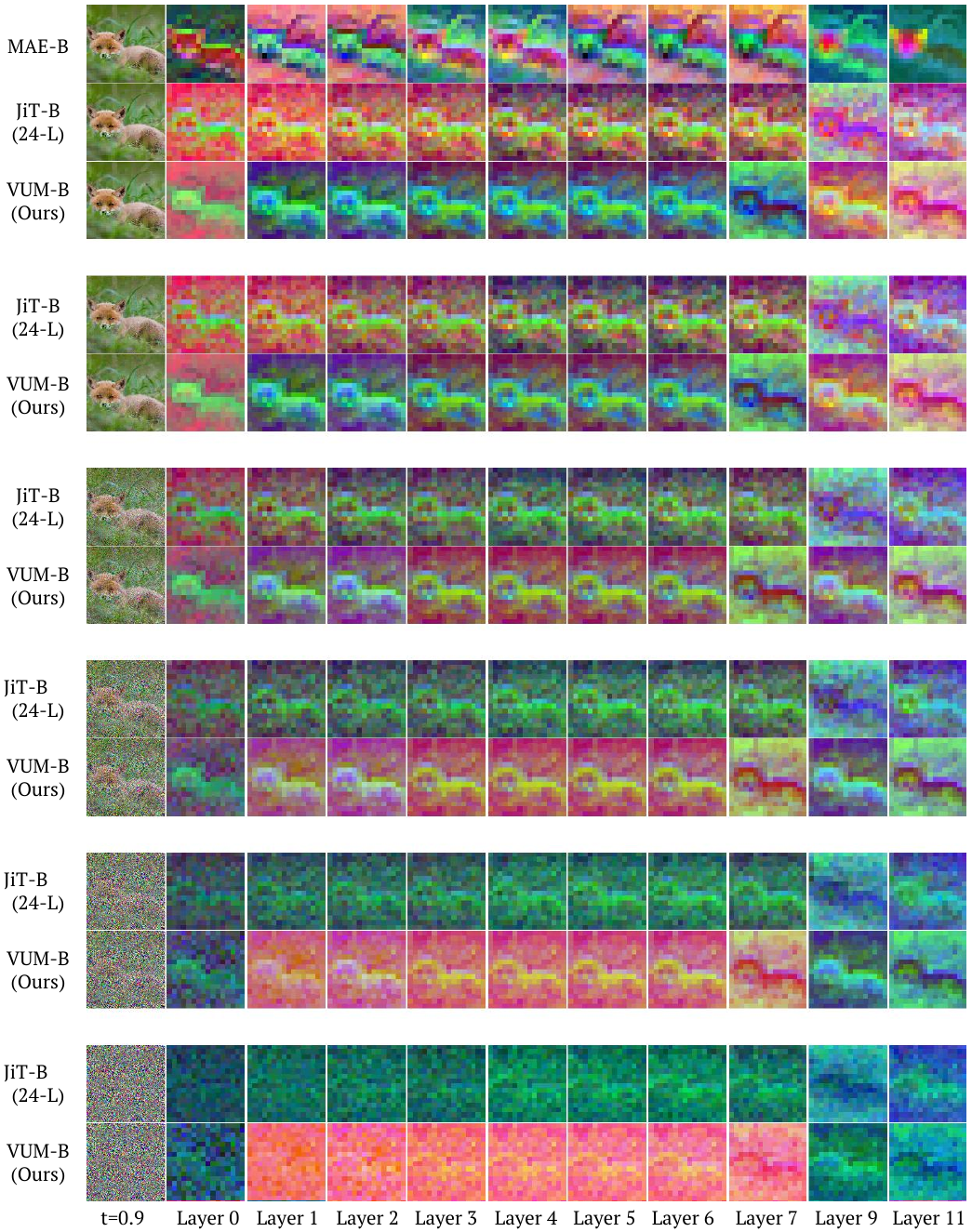}}
\caption{PCA feature map comparison among MAE-B, JiT-B (24 layers), and VUM-B. MAE-B is evaluated exclusively on clean images ($t=0$), whereas JiT-B and VUM-B are evaluated across varying noise levels $t \in \{0, 0.1, 0.3, 0.5, 0.7, 0.9\}$. For all models, features are extracted from layers $\{0, 1, 2, 3, 4, 5, 6, 7, 9, 11\}$.}
\label{fig:pca_4}
\end{center}
\end{figure*}

We present the PCA visualization of the feature maps in \cref{fig:pca_1,fig:pca_2,fig:pca_3,fig:pca_4}. We use MAE-B, JiT-B (24 layers) and our VUM-B (i.e., the three models in \cref{tab:ablations_component}). MAE-B is evaluated exclusively on clean images ($t=0$), whereas JiT-B and VUM-B are evaluated on varying noise levels $t \in \{0, 0.1, 0.3, 0.5, 0.7, 0.9\}$. For all models, the features are extracted from the layers $\{0, 1, 2, 3, 4, 5, 6, 7, 9, 11\}$. Compared to JiT-B, which produces noisy feature maps, VUM-B consistently produces cleaner and more structured representations. These discriminative hidden states directly benefit both generation quality and downstream perception performance.

\section{Qualitative Results}
\label{app:qualitative_results}

\begin{figure*}[htbp]
\begin{center}
\centerline{\includegraphics[width=1.0\linewidth]{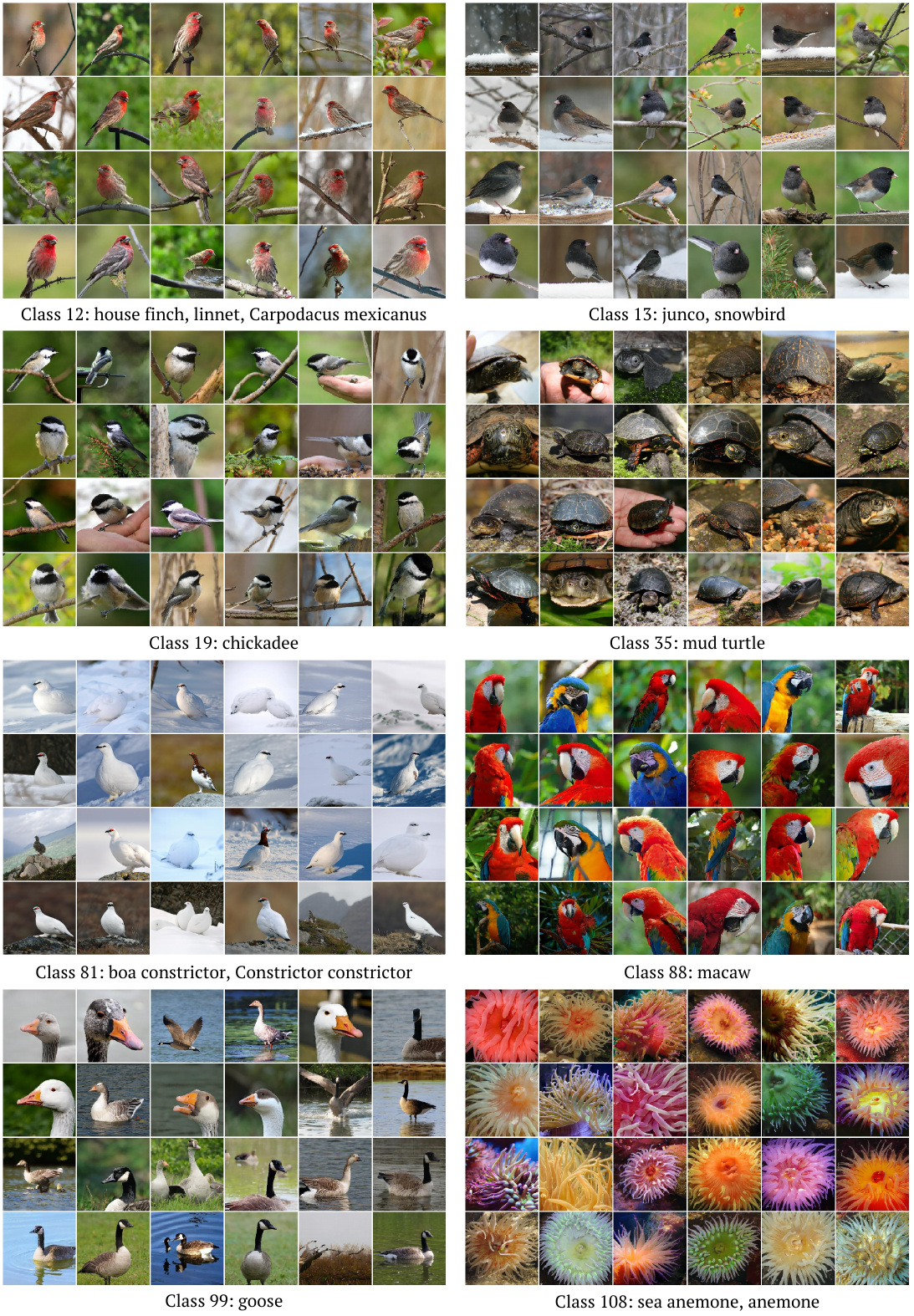}}
\caption{Uncurated, class-conditional samples on ImageNet $256\times256$ using VUM-L.}
\label{fig:vis_1}
\end{center}
\end{figure*}

\begin{figure*}[htbp]
\begin{center}
\centerline{\includegraphics[width=1.0\linewidth]{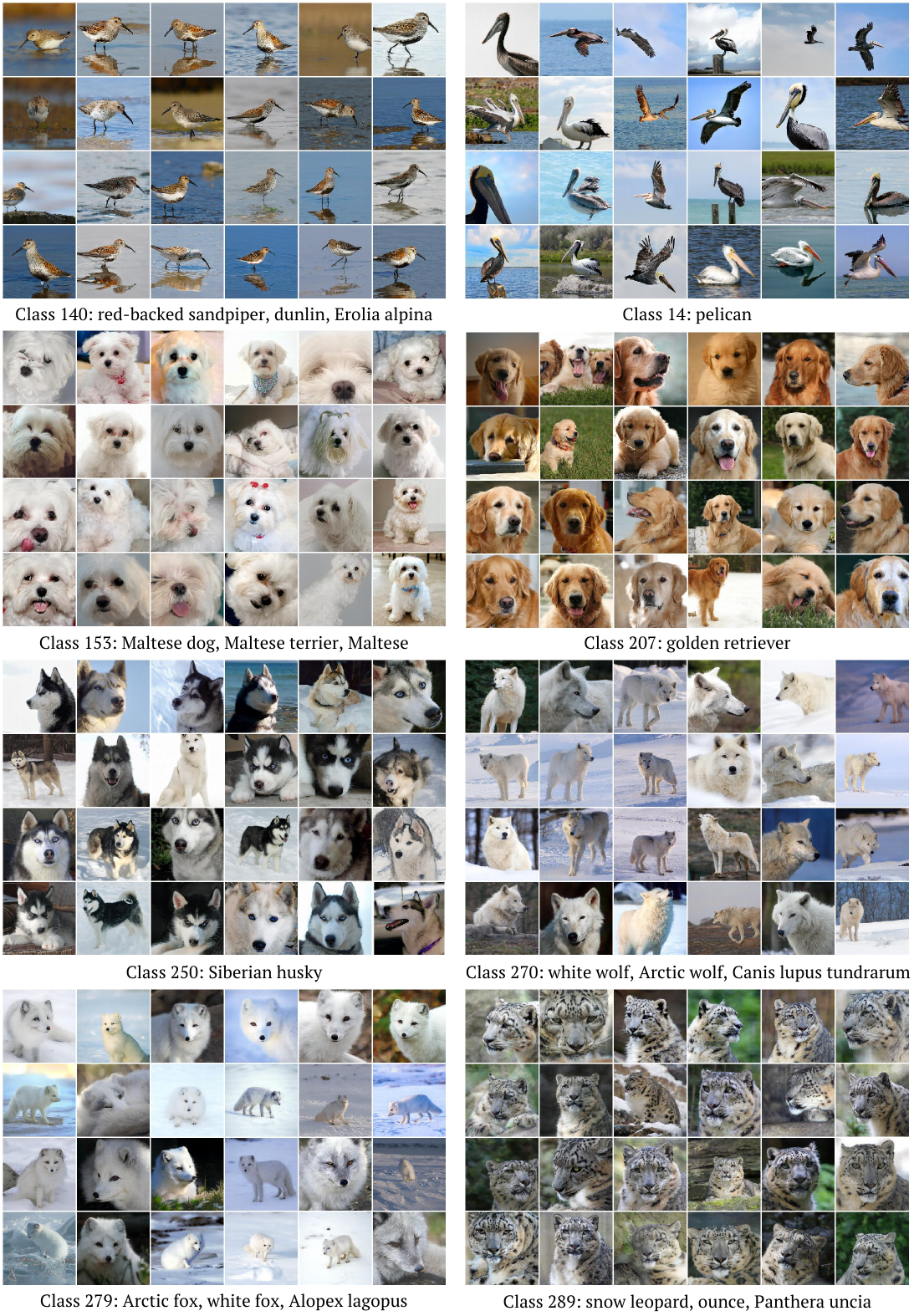}}
\caption{Uncurated, class-conditional samples on ImageNet $256\times256$ using VUM-L.}
\label{fig:vis_2}
\end{center}
\end{figure*}

\begin{figure*}[htbp]
\begin{center}
\centerline{\includegraphics[width=1.0\linewidth]{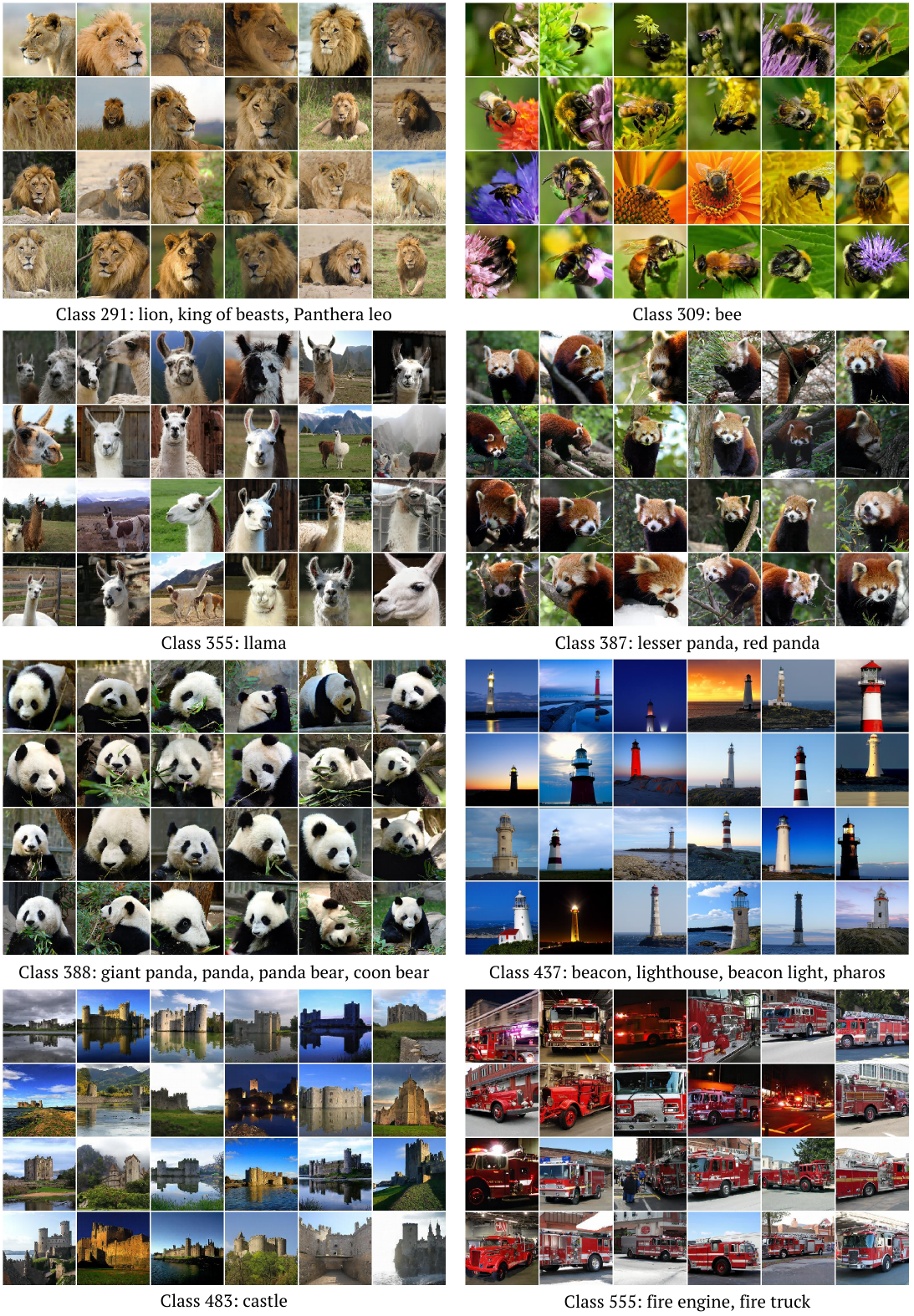}}
\caption{Uncurated, class-conditional samples on ImageNet $256\times256$ using VUM-L.}
\label{fig:vis_3}
\end{center}
\end{figure*}

\begin{figure*}[htbp]
\begin{center}
\centerline{\includegraphics[width=1.0\linewidth]{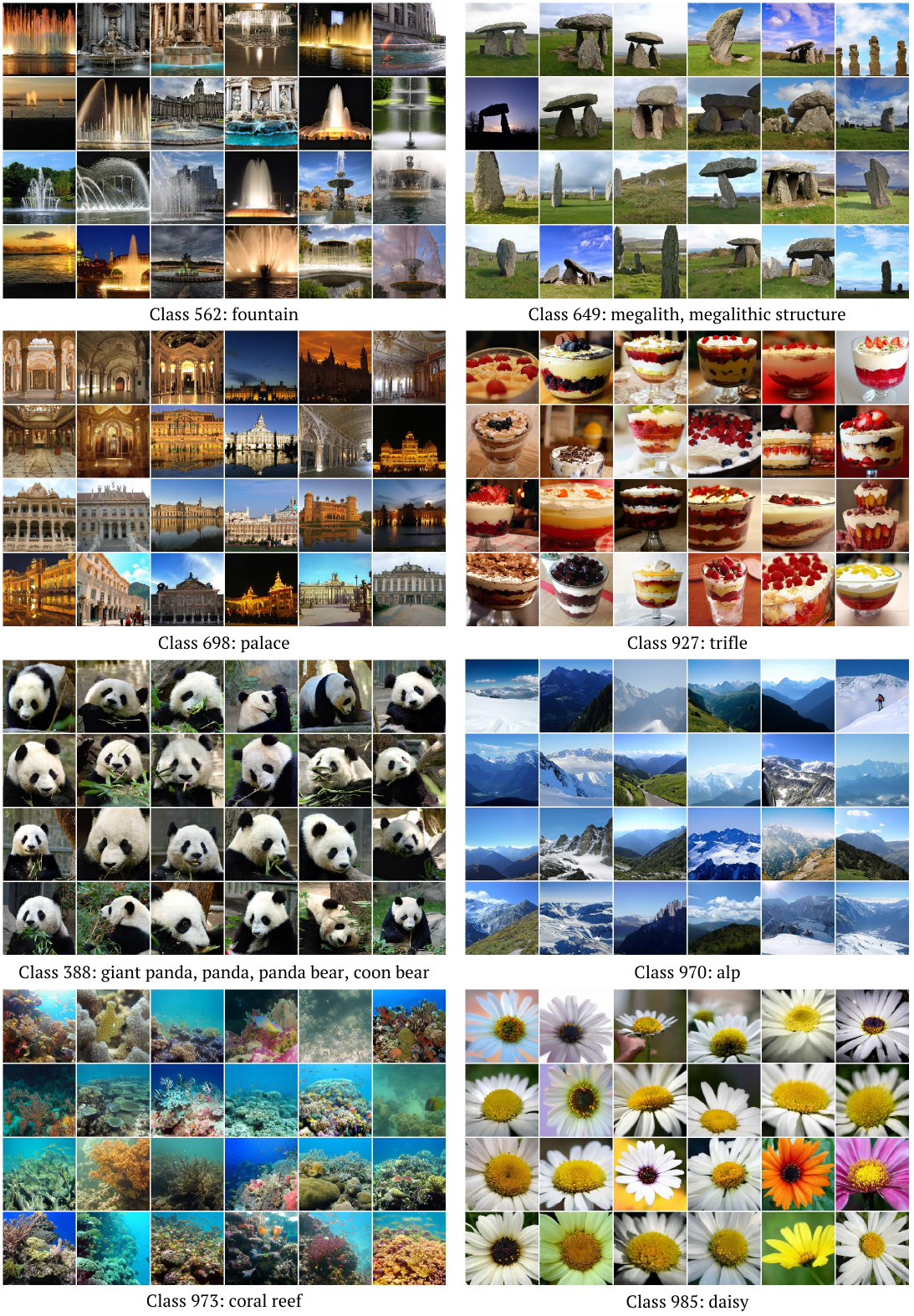}}
\caption{Uncurated, class-conditional samples on ImageNet $256\times256$ using VUM-L.}
\label{fig:vis_4}
\end{center}
\end{figure*}

We present uncurated qualitative results generated by VUM-L in \cref{fig:vis_1,fig:vis_2,fig:vis_3,fig:vis_4}. All images are synthesized using 100 ODE sampling steps with a classifier-free guidance (CFG) scale of 2.5.

\clearpage
\section*{NeurIPS Paper Checklist}

\begin{enumerate}

\item {\bf Claims}
    \item[] Question: Do the main claims made in the abstract and introduction accurately reflect the paper's contributions and scope?
    \item[] Answer: \answerYes{} 
    \item[] Justification: We provide detailed experimental proofs to support our claims.

\item {\bf Limitations}
    \item[] Question: Does the paper discuss the limitations of the work performed by the authors?
    \item[] Answer: \answerYes{} 
    \item[] Justification: We discuss the limitations in the final sections.

\item {\bf Theory assumptions and proofs}
    \item[] Question: For each theoretical result, does the paper provide the full set of assumptions and a complete (and correct) proof?
    \item[] Answer: \answerNA{} 
    \item[] Justification: This paper does not include new theorems or theoretical assumptions.

    \item {\bf Experimental result reproducibility}
    \item[] Question: Does the paper fully disclose all the information needed to reproduce the main experimental results of the paper to the extent that it affects the main claims and/or conclusions of the paper (regardless of whether the code and data are provided or not)?
    \item[] Answer: \answerYes{} 
    \item[] Justification: We provide enough information and implementation details to reproduce the experimental results.

\item {\bf Open access to data and code}
    \item[] Question: Does the paper provide open access to the data and code, with sufficient instructions to faithfully reproduce the main experimental results, as described in supplemental material?
    \item[] Answer: \answerYes{} 
    \item[] Justification: We use the public-available data only.

\item {\bf Experimental setting/details}
    \item[] Question: Does the paper specify all the training and test details (e.g., data splits, hyperparameters, how they were chosen, type of optimizer) necessary to understand the results?
    \item[] Answer: \answerYes{} 
    \item[] Justification: We provide enough training and test details.

\item {\bf Experiment statistical significance}
    \item[] Question: Does the paper report error bars suitably and correctly defined or other appropriate information about the statistical significance of the experiments?
    \item[] Answer: \answerYes{} 
    \item[] Justification: We report detailed experimental results.

\item {\bf Experiments compute resources}
    \item[] Question: For each experiment, does the paper provide sufficient information on the computer resources (type of compute workers, memory, time of execution) needed to reproduce the experiments?
    \item[] Answer: \answerYes{} 
    \item[] Justification: We provide the information of our compute resources.
    
\item {\bf Code of ethics}
    \item[] Question: Does the research conducted in the paper conform, in every respect, with the NeurIPS Code of Ethics \url{https://neurips.cc/public/EthicsGuidelines}?
    \item[] Answer: \answerYes{} 
    \item[] Justification:We respect and follow the NeurIPS Code of Ethics.

\item {\bf Broader impacts}
    \item[] Question: Does the paper discuss both potential positive societal impacts and negative societal impacts of the work performed?
    \item[] Answer: \answerNA{} 
    \item[] Justification: This paper is focused on foundation model design, which does not involve such problems.
    
\item {\bf Safeguards}
    \item[] Question: Does the paper describe safeguards that have been put in place for responsible release of data or models that have a high risk for misuse (e.g., pre-trained language models, image generators, or scraped datasets)?
    \item[] Answer: \answerNA{} 
    \item[] Justification: This paper is focused on foundation model design, which does not involve such problems.

\item {\bf Licenses for existing assets}
    \item[] Question: Are the creators or original owners of assets (e.g., code, data, models), used in the paper, properly credited and are the license and terms of use explicitly mentioned and properly respected?
    \item[] Answer: \answerYes{} 
    \item[] Justification: We respect the credit and license, terms of existing code, data, models used in our research.

\item {\bf New assets}
    \item[] Question: Are new assets introduced in the paper well documented and is the documentation provided alongside the assets?
    \item[] Answer: \answerNA{} 
    \item[] Justification: Our research is focused on the network architecture and training  paradigm design.

\item {\bf Crowdsourcing and research with human subjects}
    \item[] Question: For crowdsourcing experiments and research with human subjects, does the paper include the full text of instructions given to participants and screenshots, if applicable, as well as details about compensation (if any)? 
    \item[] Answer: \answerNA{} 
    \item[] Justification: Not involved.Not involved.

\item {\bf Institutional review board (IRB) approvals or equivalent for research with human subjects}
    \item[] Question: Does the paper describe potential risks incurred by study participants, whether such risks were disclosed to the subjects, and whether Institutional Review Board (IRB) approvals (or an equivalent approval/review based on the requirements of your country or institution) were obtained?
    \item[] Answer: \answerNA{} 
    \item[] Justification: Not involved.

\item {\bf Declaration of LLM usage}
    \item[] Question: Does the paper describe the usage of LLMs if it is an important, original, or non-standard component of the core methods in this research? Note that if the LLM is used only for writing, editing, or formatting purposes and does \emph{not} impact the core methodology, scientific rigor, or originality of the research, declaration is not required.
    \item[] Answer: \answerNA{} 
    \item[] Justification: LLM is only a tool in our research.

\end{enumerate}


\end{document}